\documentclass[runningheads]{llncs}
\usepackage[T1]{fontenc}
\usepackage{graphicx}
\usepackage{booktabs}
\usepackage[misc]{ifsym}
\newcommand{\corr}{(\Letter)}
\usepackage{mwe}
\usepackage{amsmath}
\usepackage{amssymb}
\usepackage{multirow}
\usepackage{pdfpages}
\usepackage{caption}
\usepackage{tcolorbox}
\usepackage{listings}
\usepackage{xcolor}
\usepackage{tabularx}
\usepackage{algorithm}
\usepackage{algorithmicx}
\usepackage{algpseudocode}
\usepackage{caption}

\definecolor{codegray}{gray}{0.9}
\definecolor{borderblue}{RGB}{50, 100, 200}

\begin{document}

\title{Zero-Shot Detection of LLM-Generated Code via Approximated Task Conditioning}

\titlerunning{Detection of LLM-Generated Code via Approximated Task Conditioning}

\author{Maor Ashkenazi\inst{1,2}\thanks{Equal contribution.} \corr \and
Ofir Brenner\inst{3}* \and
Tal Furman Shohet\inst{4} \and Eran Treister\inst{1,2} }

\authorrunning{M. Ashkenazi and O. Brenner et al.}

\institute{Department of Computer Science, Ben-Gurion University of the Negev \email{maorash@post.bgu.ac.il}
\and
Data Science Research Center, Ben-Gurion University of the Negev
\and
Tel-Aviv University
\and
Deep Instinct}

\tocauthor{Maor Ashkenazi, Ofir Brenner, Tal Furman Shohet, Eran Treister}
\toctitle{Zero-Shot Detection of LLM-Generated Code via Approximated Task Conditioning}

\maketitle              %

\begin{abstract}
Detecting Large Language Model (LLM)-generated code is a growing challenge with implications for security, intellectual property, and academic integrity. 
We investigate the role of conditional probability distributions in improving zero-shot LLM-generated code detection, when considering both the code and the corresponding task prompt that generated it.
Our key insight is that when evaluating the probability distribution of code tokens using an LLM, there is little difference between LLM-generated and human-written code. However, conditioning on the task reveals notable differences.
This contrasts with natural language text, where differences exist even in the unconditional distributions.
Leveraging this, we propose a novel zero-shot detection approach that approximates the original task used to generate a given code snippet and then evaluates token-level entropy under the \textit{approximated task conditioning (ATC)}. We further provide a mathematical intuition, contextualizing our method relative to previous approaches. 
ATC requires neither access to the \textit{generator LLM} nor the original task prompts, making it practical for real-world applications. To the best of our knowledge, it achieves state-of-the-art results across benchmarks and generalizes across programming languages, including Python, CPP, and Java. Our findings highlight the importance of task-level conditioning for LLM-generated code detection. 
The supplementary materials and code are available at \url{https://github.com/maorash/ATC}, including the dataset gathering implementation, to foster further research in this area.

\keywords{LLMs
\and Synthetic text detection \and Synthetic code detection}
\end{abstract}

\section{Introduction}
\label{introduction}
Large Language Models (LLMs) such as Claude \cite{anthropic2024} and GPT \cite{openai2024} have demonstrated remarkable capabilities in text generation, excelling in tasks such as summarization, translation, and creative writing. These models typically leverage the transformer architecture~\cite{vaswani2017attention} and large-scale pretraining to produce coherent text. 
However, their broad adoption has raised concerns about misinformation and other ethical challenges~\cite{zellers2019defending,bommasani2021opportunities,jawahar2020automatic}, highlighting the need for robust detection methods. %
More recently, LLMs have shown impressive proficiency in code generation, with models like CodeLlama~\cite{roziere2023code}, and StarCoder~\cite{Tunstall2023starchat-alpha} producing functional code snippets. These advancements transformed software development by automating repetitive coding tasks, assisting with debugging, and even generating novel solutions from high-level descriptions. Furthermore, AI coding agents and integrated development tools have transformed modern workflows by embedding generation capabilities directly into the programming process. Although these advances improved productivity, they also introduce concerns related to security, intellectual property, and academic integrity. As a result, distinguishing LLM-generated code from human-written code is crucial for mitigating potential risks.
While significant progress has been made in detecting LLM-generated natural language text, identifying LLM-generated code remains a challenging problem. Prior research attributes this difficulty to the structured nature of code, which results in lower predictive token entropy compared to natural language~\cite{ye2024uncovering}. Unlike natural language, where lexical choices and sentence structures vary widely, programming languages impose strict syntactic and semantic rules, making token probability distributions less informative for detection. 
In our research, we take a different approach by analyzing the role of task conditioning in improving detection.
To investigate this, we conduct an initial experiment comparing LLM-generated and human-written content across natural language and code. We use two datasets: MBPP~\cite{austin2021program}, which consists of programming tasks and code snippets, and WritingPrompts~\cite{huang2024gpt}, which contains natural language stories. We use the entire test set from MBPP and sample an equivalent amount of texts from WritingPrompts, generating responses using CodeLlama for MBPP and LLaMA 3.1~\cite{meta_llama3.1} for WritingPrompts. To avoid unwanted effects of response lengths, responses shorter than 200 characters are discarded, while longer ones are truncated. For each response, we compute mean token entropy using the same model that generated it, under two settings: (1) unconditional sampling and (2) sampling conditioned on the original task. Extended details on the initial experiment are provided in Appendix A.
The results in Figure~\ref{fig:observation} reveal a clear pattern. Without task conditioning, the entropy distributions of human-written and LLM-generated code overlap significantly, making detection difficult. In contrast, human-written and LLM-generated text from WritingPrompts exhibit greater separability even without task conditioning. When introducing task conditioning, both datasets show improved distinguishability between human-written and LLM-generated content. This finding suggests that detecting LLM-generated code without access to the task prompt is inherently challenging, but incorporating task-level context can enhance detection accuracy. Intuitively, conditioning on the task provides the LLM with specific context for the expected code, allowing the conditional distribution to primarily focus on coding style, instead of code purpose.

\begin{figure}[t]
    \centering
    \begin{minipage}{0.48\textwidth}
        \centering
        \includegraphics[width=1\linewidth, trim=0.2cm 0.4cm 0.2cm 0.6cm, clip]{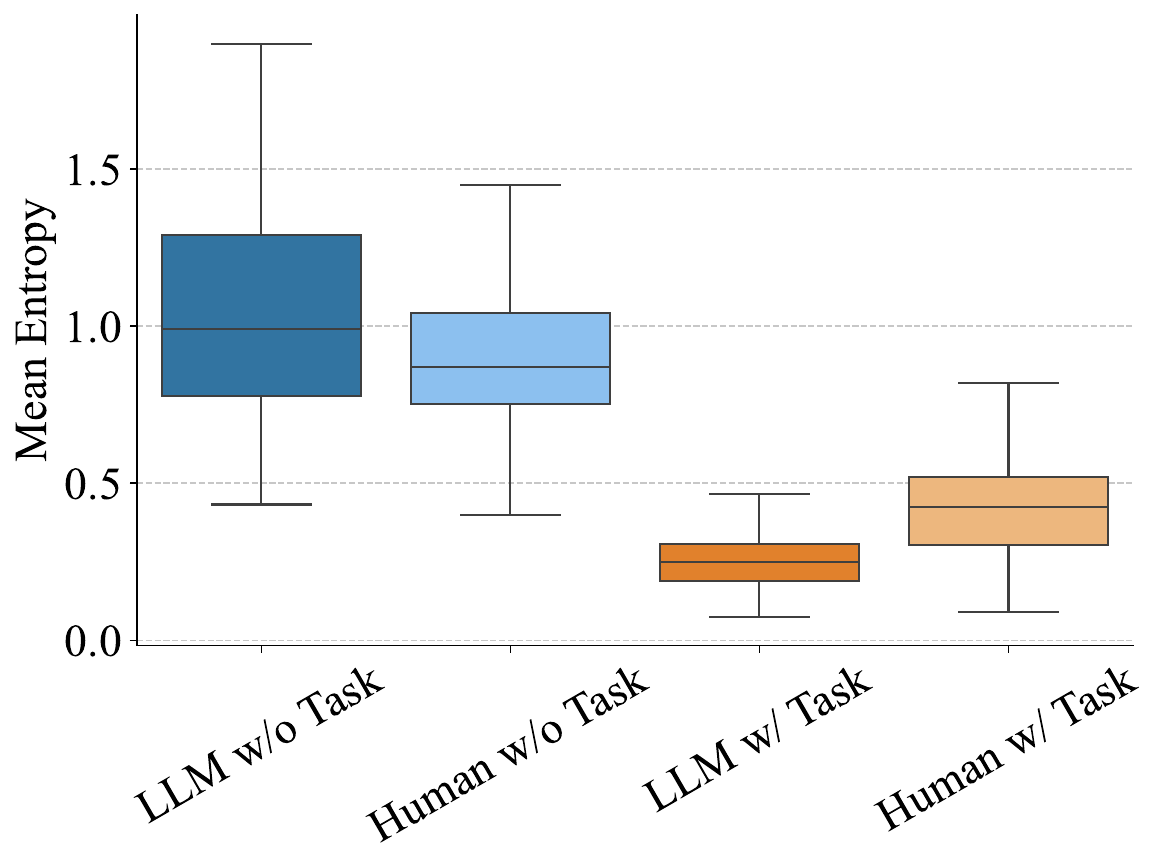}
    \end{minipage}
    \hfill
    \begin{minipage}{0.48\textwidth}
        \centering
        \includegraphics[width=1\linewidth, trim=0.2cm 0.4cm 0.2cm 0.6cm, clip]{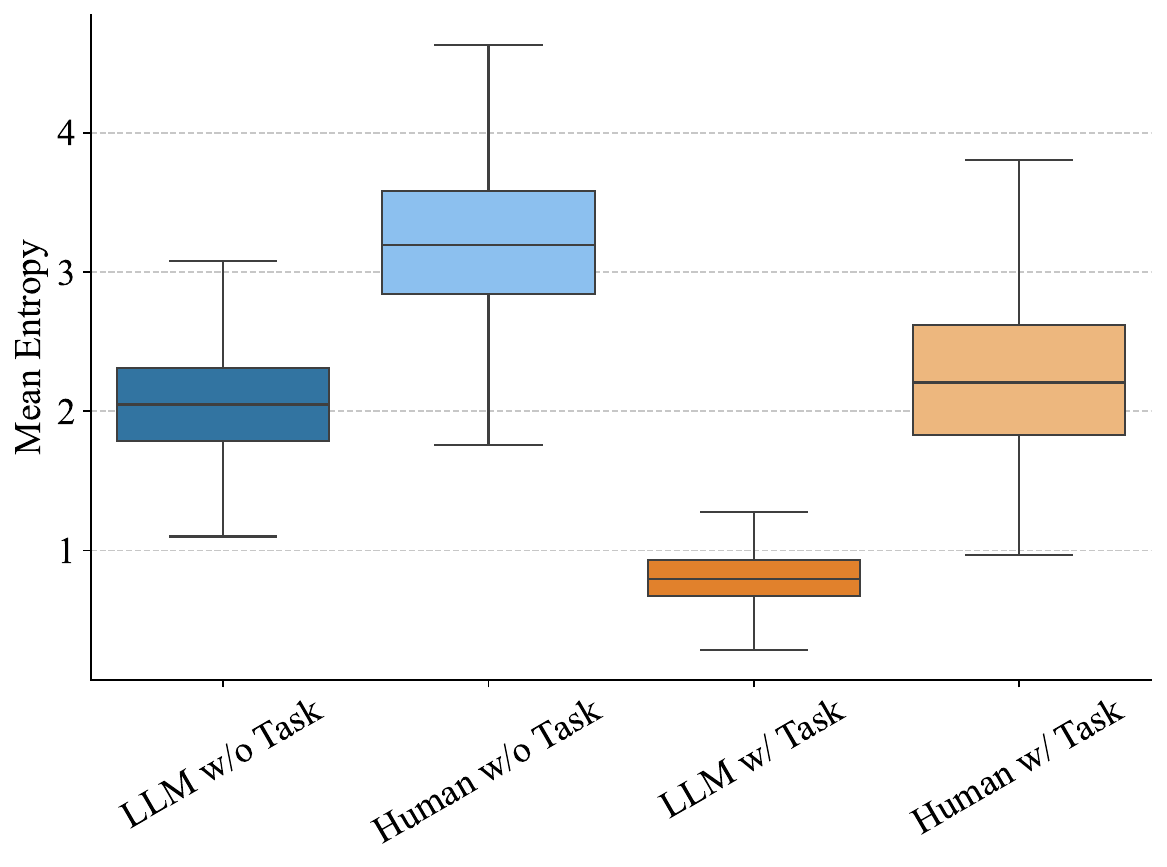} 
    \end{minipage}
    \caption{Box plot of mean token entropy values for human and LLM-generated texts. MBPP (code) on the left, WritingPrompts (natural language) on the right.
    }
    \label{fig:observation}
\end{figure}

Building on this, we introduce \textit{Approximated Task Conditioning (ATC)}, a novel zero-shot detection approach that approximates a task from a given code snippet and evaluates token entropy under its conditioning. \textit{ATC} does not require access to the original task but instead approximates it in a relatively lightweight manner, making it practical for real-world scenarios where task information is unavailable. 
We summarize our contributions as follows:
\begin{itemize}
    \item We propose \textit{ATC}, a novel zero-shot approach for detecting LLM-generated code that achieves state-of-the-art (SOTA) detection results.
    \item We establish a connection between \textit{ATC} and prior detection methods.%
    \item We conduct extensive experiments demonstrating the robustness of \textit{ATC}.
    \item We release our code, including dataset gathering implementation, fostering collaboration and further research in the field.
\end{itemize}

\section{Related Work}
\textbf{Detecting LLM-Generated Text}
As LLMs become more widespread, distinguishing between human and LLM-generated text becomes increasingly important to combat issues like fake news and plagiarism~\cite{zellers2019defending,bommasani2021opportunities}. Detection methods can generally be divided into two categories: supervised learning and zero-shot approaches.
Supervised learning techniques involve training models to differentiate between human-written and LLM-generated text~\cite{zhong2020neural,yu2023gpt,mitrovic2023chatgpt}. These methods often struggle with generalization, as they tend to overfit to specific datasets or the LLMs used for text generation~\cite{bakhtin2019real,pu2023deepfake,mitchell2023detectgpt}. In contrast, zero-shot approaches have shown greater robustness, focusing on analyzing token distribution patterns like entropy~\cite{thomas2008detecting}, likelihood~\cite{gehrmann2019gltr}, and ranks~\cite{su2023detectllm} to detect signs of LLM-generated text. Perturbation-based methods like DetectGPT~\cite{mitchell2023detectgpt} and NPR~\cite{su2023detectllm} perturb text and analyze differences between the original and perturbed texts. Similarly, \cite{yangdna} generates completions and compares their similarity to the original text.
These methods have higher computational costs due to repeated iterations, whereas FastDetectGPT~\cite{bao2023fast} improves efficiency by optimizing the perturbations.

\bigskip
\noindent\textbf{Detecting LLM-Generated Code}  
Text-based detectors often struggle with code due to structural differences from natural language~\cite{lee2024wrote,ye2024uncovering}. This has led to the development of specialized detection methods.  
\cite{yang2023zero} adapts DetectGPT with fill-in-the-middle masking which replaces entire lines of code, while \cite{shi2025detectcodegpt} uses stylistic patterns such as whitespace changes in addition to preserving code correctness after perturbations. \cite{xu2024detecting} proposes a method using targeted perturbations and a fine-tuned CodeBERT model. %
\cite{xu2024investigating} conducted a large-scale evaluation of detection methods, finding that while some generalize well, they struggle often with high-level languages and short code snippets. 
The most recent and, to our knowledge, state-of-the-art (SOTA) method is \cite{ye2024uncovering}, which generates code variants via multiple rewriting prompts and measures their similarity. However, this requires numerous rewrites and training a code similarity model.  
In contrast, our method requires less iterations, achieving superior performance with a single LLM prompt and no additional training. 
Meanwhile, data collection efforts, such as \cite{demirok2024aigcodeset}, are providing benchmarks for future detection research.

\section{Method}
We consider the problem of zero-shot LLM-generated code detection. Given a code snippet $x$, we wish to determine whether it was generated by an LLM, or written by a human. 
We use an open-source \textit{detector LLM} to evaluate token probability distributions, and do not assume access or knowledge of the \textit{generator LLM}, used for generating the code. In addition, our approach is \textit{zero-shot}, meaning it does not require labeled training data nor involves any training steps, resulting in robust results across \textit{generator LLMs} and programming languages.

\subsection{Approximated Task Conditioning (ATC)}
\begin{figure}[t]
    \centering
    \includegraphics[width=1\linewidth, trim=0cm 6.9cm 8.1cm 0cm, clip]{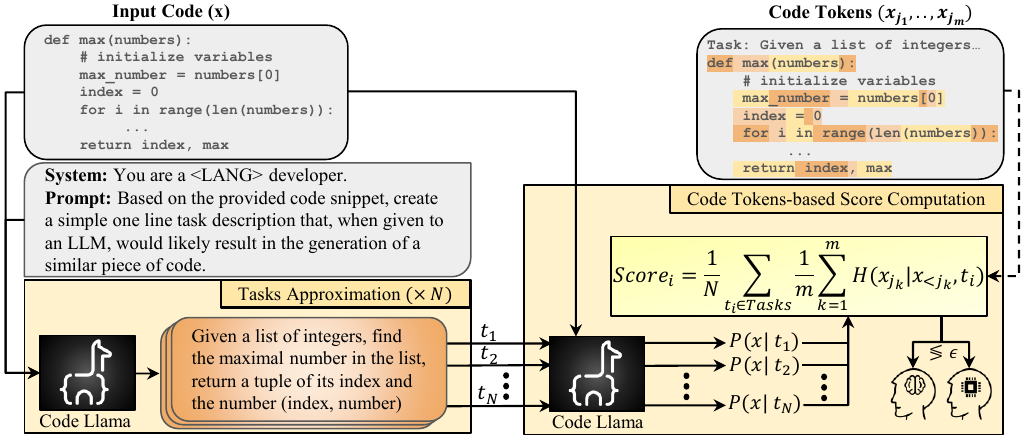}
    \caption{ 
    \textbf{Overview of \textit{ATC}.} Given an input code snippet $x$, we (1) query the \textit{detector LLM} (CodeLlama) with a fixed prompt to generate task descriptions \(t_1,..,t_N\) for which $x$ might be a valid solution, and (2) compute the conditional entropy of the input code tokens given each approximated task. Given the probability distribution \(P(x \mid t_i)\) for each task \(t_i\), the final score is obtained by averaging the mean token entropy \textit{only on code tokens} (see top-right). Low entropy scores indicate higher confidence that the code was LLM-generated.
    }
    \label{fig:main}
\end{figure}

Here we elaborate on our LLM-generated code detection method (\textit{ATC}), consisting of two main steps. First, we approximate one or more tasks for the input code snippet by prompting an LLM, which we term the \textit{detector LLM}. Next, we calculate the score for the given code sample by computing the mean token entropy of the conditional distribution on each of the approximated tasks. When computing the score, we only consider code tokens, i.e., we ignore the task and comment tokens' entropy.
We use the same \textit{detector LLM} for token sampling for consistency. Our approach is detailed in Algorithm~\ref{alg:atc}. A visualization of the pipeline and prompt used for task approximation is in Figure~\ref{fig:main}, alongside an example input, with details and connection to previous approaches below.

\begin{algorithm}[h]
\caption{Approximated Task Conditioning (ATC)}
\label{alg:atc}
\begin{algorithmic}[1]
\Require $x$: code, $N$: number of approximated tasks, $\mathcal{G}$: Detector LLM (vocabulary $\mathcal{V}$), $\epsilon$: threshold
\Ensure Decision: \texttt{LLM-Generated} or \texttt{Human-Written}
\State Query $\mathcal{G}$ with $x$ and the prompt in Figure~\ref{fig:main} $N$ times to generate task descriptions $t_1, \dots, t_N$. \textbf{ // Task Approximation}
\For{$i = 1$ to $N$} \textbf{ // Code Tokens-based Score Computation}
    \State Concatenate the texts $t_i$ and $x$.
    \State Perform a forward pass through $\mathcal{G}$ to get the conditional distribution $P(x \mid t_i)$.
    \State Get the subsequence of $m$ \textbf{code tokens}, excluding comments: $(x_{j_1},.., x_{j_m}) \subseteq x$.
    \State Compute the score for task $i$ by calculating mean code token entropy:
    \Statex \;\;\;\;
    $
    Score_i = -\frac{1}{m} \sum_{k=1}^{m} \sum_{v \in \mathcal{V}} P(v \mid x_{<j_k}, t_i) \log P(v \mid x_{<j_k}, t_i)
    $
\EndFor
\If{$\frac{1}{N} \sum_{i=1}^{N} Score_i > \epsilon$}  
    \State \Return \texttt{Human-Written}
\Else{} \Return \texttt{LLM-Generated}
\EndIf
\end{algorithmic}
\end{algorithm}

\bigskip
\noindent\textbf{Choosing the Detector LLM}
We consider a relatively small and open-source LLM, CodeLlama13b, alongside its smaller counterpart, CodeLlama7b, as opposed to previous methods which relied on proprietary models available via APIs to achieve good results. We show that using CodeLlama7b is enough to surpass previous methods, and that using a larger model improves performance. Furthermore, we show that these relatively small \textit{detector LLMs} achieve robust performance across various \textit{generator LLMs} and tasks. %

\bigskip
\noindent\textbf{Task Approximation}
This step is performed by querying the \textit{detector LLM} with a fixed prompt, asking it to generate a task that, when given to an LLM, would likely produce a similar code snippet to $x$. The full prompt is presented in Figure~\ref{fig:main}. We use $top_p = 0.95$ and a temperature of $0.7$ for sampling, similar to how we generate the code solutions for the experiments. Setting $top_p$ will limit sampling to the most probable tokens whose cumulative probability reaches 0.95, and the temperature controls the randomness of the sampling. Additional details are in Appendix B. While a single approximated task already outperforms current SOTA, our experiments show that generating multiple tasks and averaging their corresponding scores further improves performance. In most experiments, we use $N = 1, 2, 4$ approximated tasks, where $N$ is a hyperparameter.

\bigskip
\noindent\textbf{Code Tokens-based Score Computation}
Given a code snippet $x$, we find the $m$ code tokens, excluding comments $(x_{j_1},..,x_{j_m}) \subseteq x$. Next, using an approximated task $t$, we compute the mean token entropy conditioned on $t$; 
\begin{equation}
\label{equation:score}
\frac{1}{m} \sum_{k=1}^{m} H(x_{j_k} \mid x_{<j_k}, t) = -\frac{1}{m} \sum_{k=1}^{m} \sum_{v \in \mathcal{V}} P(v \mid x_{<j_k}, t) \log P(v \mid x_{<j_k}, t)
\end{equation}
where $\mathcal{V}$ is the set of all possible tokens in the vocabulary. 
Intuitively, the entropy of the distribution should be lower (i.e., the model should be more confident) for LLM-generated code, as such code is more likely to align with the \textit{detector LLM}'s learned distribution. To obtain the final score, we average the mean token entropy over all approximated tasks $t_1,..,t_N$.
\textit{ATC} can be integrated with other baseline approaches, such as computing the mean log likelihood or analyzing mean token ranks. However, we find that our method is most effective when used alongside entropy estimation, as seen in Section~\ref{ablation:scoring_methods}.

\bigskip
\noindent\textbf{Handling Comment Tokens}
Both human-written and LLM-generated code often include comments, such as inline comments and docstrings. We analyze the tendency to add comments in Appendix C.
Due to the autoregressive nature of code generation, preceding comment tokens can significantly influence the conditional distribution of subsequent tokens. If comments accurately describe relevant parts of the code, they effectively act as a more localized and specific task within the code snippet. By treating comments this way, we aim to capture finer-grained task information, which can further aid in distinguishing between human-written and LLM-generated code.
Thus, we handle comment tokens similarly to the task tokens, i.e. we exclude them from the token entropy calculation. However, \textit{comments remain part of the input when modeling the conditional distribution}. While it is reasonable to assume that comments accurately describe relevant code, we also test robustness against adversarial modifications by evaluating the impact of removing comments before scoring (see Section~\ref{section:removing_comments}).

\subsection{Intuition and Connection to Previous Methods}
Here we build an intuition, relating our method to zero-shot baseline methods and to ~\cite{ye2024uncovering} which we consider the current SOTA.
Our observation in Figure~\ref{fig:observation} emphasizes that the predictive entropy of code tokens cannot be clearly distinguished when task conditioning is unavailable, i.e. the unconditional distributions between human-written and LLM-generated code are less separable than the conditional distributions. 
Baseline methods (e.g. log likelihood~\cite{gehrmann2019gltr} or token entropy~\cite{thomas2008detecting}) assess the certainty of sampling a code snippet $x$ under the LLM's unconditional distribution $P(x)$, since they do not have access to the task. Assuming a latent task variable $t^* \sim P_t$, the unconditional distribution of code is given by
\begin{equation}
P(x) = \int P(x \mid t^*) P_t(t^*) \, dt^*.    
\end{equation}

In \cite{ye2024uncovering}, the model is prompted solely with the code snippet $x$ and generates a rewrite $x'$ from the conditional distribution $ P(x' \mid x) $.
Next, the similarity between $x'$ and $x$ is assessed, and high similarity leads to high confidence that $x$ is LLM-generated.
We argue that our method shares similarities with \cite{ye2024uncovering}, particularly in how both approaches approximate the latent task $t^*$. Specifically, we propose that sampling from $P(x' \mid x)$ serves as an approximation to $P(x' \mid t^*)$. This is based on an assumption that conditioning on $x$ might be similar to conditioning on $t^*$, as the original code $x$ inherently carries implicit information about the underlying task $t^*$. Thus, one may argue an approximation of:
\begin{equation}
P(x' \mid x) \approx P(x' \mid t^*).    
\end{equation}

Intuitively, if a model is asked to find the maximal number in a list, its response will likely resemble the output it produces when asked to rewrite an existing snippet that does so. Sampling multiple rewrites and measuring their similarity,
\begin{equation}
\mathbb{E}_{x' \sim P(\cdot \mid x)}\left[S(x,x')\right],    
\end{equation}
should be correlated with the probability of $x$ being drawn from $P(x' \mid t^*)$.

In contrast, our approach explicitly approximates the task itself $t \approx t^*$, by prompting the detector LLM. We then calculate the token entropy of $x$ under $P(x \mid t)$. This two-step process, inspired by Chain-of-Thought principles, empirically provides a more interpretable and accurate estimation of the conditional distribution:
\begin{equation}
P(x \mid t) \approx P(x \mid t^*).
\end{equation}

Finally, \cite{ye2024uncovering} employs multiple rewrites, which may suggest that their approximation requires multiple iterations to refine the conditional distribution estimate. In contrast, \textit{ATC} achieves better results with a single approximation.%

\section{Experiments}
This section details our experimental setup, covering datasets, generation models, and baseline methods. We then present the main Python results, assess the impact of comment removal as a pre-processing step, compare the approximated tasks to original ones, and analyze different task approximation prompts. We also examine robustness across factors like decoding strategies, different programming languages, and code length. Finally, we conduct relevant ablation experiments. 
In all experiments, we use AUROC (Area Under the Receiver Operating Characteristic curve) to evaluate performance, following previous works~\cite{ye2024uncovering,shi2025detectcodegpt,mitchell2023detectgpt}.
Additionally, we explore alternative metrics in Section~\ref{section:real_world}.

\subsection{Experimental Setup}
\label{section:experimental_setup}
\noindent\textbf{Datasets}
To compare with the current SOTA method \cite{ye2024uncovering}, we evaluate our approach using two widely recognized benchmarks for Python code generation: APPS~\cite{hendrycks2021measuring} and MBPP~\cite{austin2021program}. To the best of our knowledge these are the most appropriate benchmarks for our task. Notably, APPS includes solutions written by a wide range of users, leading to diverse coding styles that better reflect real-world variability. APPS contains 5,000 test instances, each consisting of a problem description and corresponding solutions. After applying the data sanitization pipeline from \cite{ye2024uncovering}, we are left with 3,765 instances. Unlike ~\cite{ye2024uncovering}, which randomly sampled 1,500 instances of the test set, we use the entire APPS test set for a more comprehensive evaluation. While differences in dataset size and sampling procedures may limit direct comparisons, this approach was necessary due to the lack of code or detailed information regarding their sampling methodology. For each instance, we select the first human solution and generate corresponding LLM outputs using each \textit{generator LLM} (detailed below). MBPP consists of Python programming problems designed for entry-level programmers, with 500 test instances. As with APPS, we use the full test set, generating LLM solutions using each generator LLM. We focus on Python due to its widespread use, readability, and versatility in code generation tasks. In both datasets we exclude the training data due to potential overlap with LLM training corpora.

\bigskip
\noindent\textbf{Generation Models}
We use a variety of open-source and proprietary models for code generation. For proprietary models, we use GPT-3.5-Turbo, GPT-4o-mini~\cite{openai2024}, and Claude3-haiku~\cite{anthropic2024}, which were selected due to their popularity among developers~\cite{stackoverflow2024survey}. For open-source models, we use Starchat-Alpha~\cite{Tunstall2023starchat-alpha}, CodeLlama-7B \& CodeLlama-13B~\cite{roziere2023code}, and CodeGemma-7B~\cite{team2024codegemma}. These models were chosen for their widespread adoption in the open-source community. We specifically use GPT-3.5-Turbo, StarChat, and CodeLlama-13B to allow direct comparison with previous methods.
For each test instance, code is generated independently by every model, resulting in a separate set of generations. We generate solutions following the schema described in~\cite{ye2024uncovering}, using Chain-of-Thought (CoT) prompting and setting $top_p = 0.95$ and the temperature to $0.7$, sampling until the EOS token is reached. To ensure clean extraction, the prompt instructs the model to output the final solution between markup tags for simple parsing. Due to space constraints, results are reported using model name abbreviations.

\bigskip
\noindent\textbf{Baselines}
We compare our method against several existing detection methods. First, we consider methods that estimate properties of a code sample's probability distribution using a surrogate model. This includes mean $log P(x)$ \cite{gehrmann2019gltr}, LogRank and Entropy \cite{mitchell2023detectgpt}, and LRR \cite{su2023detectllm}, which combines the first two methods.
Next, we look at perturbation-based methods, which estimate a code sample's properties under small modifications. We begin with DetectGPT \cite{mitchell2023detectgpt} and NPR \cite{su2023detectllm}, adapting them to code by replacing T5\textsubscript{large} \cite{raffel2020exploring} with CodeT5\textsubscript{large} \cite{wang-etal-2021-codet5}, which, as suggested by \cite{ye2024uncovering}, improves performance. We also consider DetectCodeGPT \cite{shi2025detectcodegpt}, which builds on previous methods by applying stylistic transformations and code correctness enforcement to the perturbations. Additionally, we include results from DetectGPT4Code \cite{yang2023zero} on Java, while omitting Python comparisons as their APPS subset covered about 3.5\% of the entire test set. Although we attempted to reproduce their fill-in-middle masking strategy, we observed a decrease in performance. We do not include \cite{xu2024detecting} in our comparisons since the authors did not release code or sufficient details to recreate their datasets. 
As a supervised baseline, we include OpenAI's RoBERTa text detector \cite{openai2024}.
Finally, we compare to \cite{ye2024uncovering}, which we consider to be the current SOTA, as it consistently outperforms prior methods across multiple settings, including CPP. However, for MBPP, \cite{shi2025detectcodegpt} achieves better results for one of the \textit{generator LLMs}. 
As \cite{ye2024uncovering} subsampled their data ($\sim 40\%$) and code generation involves inherent randomness, we report their results as-is, acknowledging potential variability in direct comparisons due to the lack of code implementation and random seed details.

\begin{table}[t]
\centering
\caption{Results on MBPP.}
\label{tab:mbpp_main}
\begin{tabular}{lccccccccc}
\toprule
Generator & CLlama7b & CLlama13b & Gemma & Starchat & Claude & GPT3.5 & GPT4om & Avg. \\
\midrule
OpenAI\textsubscript{large} & 52.58 & 49.85 & 26.86 & 39.40 & 31.54 & 49.90 & 40.86 & 41.57 \\
Ye~\cite{ye2024uncovering} & - & 86.21 & - & 79.23 & - & 86.23 & - & - \\
\midrule
\multicolumn{9}{c}{Using CodeLlama7b as Detector LLM} \\
\midrule
DetectGPT & 52.07 & 54.10 & 76.82 & 74.52 & 74.31 & 59.76 & 60.26 & 64.55 \\
NPR~\cite{su2023detectllm} & 78.20 & 76.09 & 71.27 & 78.34 & 80.95 & 73.46 & 73.37 & 75.95 \\
Shi~\cite{shi2025detectcodegpt} & 69.25 & 70.48 & 86.46 & 83.06 & 84.10 & 71.32 & 73.84 & 76.93 \\
\midrule
Entropy & 45.68 & 48.28 & 64.47 & 59.58 & 62.38 & 55.16 & 54.75 & 55.76 \\
$log P(x)$ & 68.37 & 69.79 & 82.78 & 77.85 & 80.72 & 72.95 & 73.29 & 75.11 \\
LogRank & 62.29 & 64.05 & 81.56 & 75.57 & 77.77 & 66.91 & 68.18 & 70.90 \\
LRR~\cite{su2023detectllm} & 31.20 & 33.67 & 65.80 & 56.57 & 54.29 & 30.72 & 39.41 & 44.52 \\
\midrule
$ATC_{N = 1}$ & 92.82 & 92.62 & 91.20 & 91.18 & 93.82 & 90.47 & 91.23 & 91.91 \\
$ATC_{N = 2}$ & 94.06 & 93.60 & 92.67 & 92.10 & 94.85 & 91.82 & 93.02 & 93.16 \\
$ATC_{N = 4}$ & 94.36 & 94.25 & \textbf{92.76} & 92.44 & 95.28 & 92.16 & 93.44 & 93.53 \\
\midrule
\multicolumn{9}{c}{Using CodeLlama13b as Detector LLM} \\
\midrule
Entropy & 46.58 & 50.00 & 63.40 & 58.72 & 61.78 & 56.56 & 55.35 & 56.06 \\
$log P(x)$ & 68.15 & 71.83 & 82.67 & 77.66 & 81.12 & 74.07 & 74.19 & 75.67 \\
LogRank & 63.45 & 67.56 & 82.27 & 76.25 & 78.91 & 69.21 & 69.86 & 72.50 \\
LRR & 37.64 & 41.72 & 70.84 & 59.43 & 59.50 & 37.48 & 41.83 & 49.78 \\
\midrule
$ATC_{N = 1}$ & 93.56 & 94.45 & 88.78 & 91.23 & 93.66 & 90.62 & 92.46 & 92.11 \\
$ATC_{N = 2}$ & 94.88 & 95.64 & 90.75 & 92.14 & 95.11 & 91.77 & 93.51 & 93.40 \\
$ATC_{N = 4}$ & \textbf{95.94} & \textbf{96.18} & 91.94 & \textbf{92.74} & \textbf{95.79} & \textbf{92.62} & \textbf{94.37} & \textbf{94.22} \\
\bottomrule
\end{tabular}
\end{table}

\subsection{Main Results}
Tables~\ref{tab:mbpp_main} and~\ref{tab:apps_main} present the results on MBPP and APPS, respectively. Our method outperforms all baselines across both datasets, consistently delivering significant improvements. It shows a clear advantage over perturbation-based methods, even when adapted to the code domain, as well as over \cite{ye2024uncovering}, which we consider the current SOTA.
Our method remains robust, maintaining high performance across a wide range of \textit{generator LLMs}, from smaller models like CodeLlama-7B to larger proprietary models like GPT-3.5.
Notably, our approach achieves superior detection performance with just a \textbf{single approximated task} ($N = 1$), whereas \cite{ye2024uncovering} relied on \textbf{eight different prompts}. For a fair comparison, setting $N = 4$ further enhances performance, yielding a mean AUROC of \textbf{94.22 on MBPP and 93.82 on APPS} when using CodeLlama-13B as the detector LLM.
The effectiveness of our method stems from the key observation made in Section~\ref{introduction}: the predictive entropy of code tokens differs significantly when sampling from the conditional distribution (i.e., conditioned on the task). 
Our method effectively approximates the task, allowing for highly accurate detection.

\begin{table}[t]
\centering
\caption{Results on APPS.}
\label{tab:apps_main}
\begin{tabular}{lccccccccc}
\toprule
Generator & CLlama7b & CLlama13b & Gemma & Starchat & Claude & GPT3.5 & GPT4om & Avg. \\
\midrule
OpenAI\textsubscript{large} & 61.11 & 59.24 & 49.01 & 49.76 & 49.84 & 47.33 & 37.43 & 50.53 \\
Ye~\cite{ye2024uncovering} & - & 87.77 & - & 82.48 & - & 83.25 & - & - \\
\midrule
\multicolumn{9}{c}{Using CodeLlama7b as Detector LLM} \\
\midrule
DetectGPT & 55.88 & 53.54 & 56.20 & 51.68 & 59.07 & 46.26 & 61.81 & 54.92 \\
NPR~\cite{su2023detectllm} & 62.12 & 60.20 & 59.08 & 55.85 & 68.21 & 53.32 & 60.75 & 59.93 \\
Shi~\cite{shi2025detectcodegpt} & 79.11 & 76.97 & 75.44 & 70.70 & 75.00 & 65.00 & 65.48 & 72.53 \\
\midrule
Entropy & 47.44 & 46.50 & 56.04 & 46.60 & 63.32 & 53.35 & 49.5 & 51.82 \\
$log P(x)$ & 67.94 & 66.49 & 72.87 & 60.84 & 73.77 & 63.91 & 54.47 & 65.75 \\
LogRank & 64.82 & 62.14 & 67.60 & 57.91 & 66.91 & 58.46 & 48.85 & 60.95 \\
LRR~\cite{su2023detectllm} & 46.75 & 40.25 & 39.45 & 42.66 & 31.89 & 34.20 & 29.54 & 37.82 \\
\midrule
$ATC_{N = 1}$ & 92.28 & 93.30 & 91.05 & 88.07 & 92.52 & 86.22 & 87.42 & 90.12 \\
$ATC_{N = 2}$ & 93.79 & 94.66 & 92.60 & 89.23 & 94.04 & 88.09 & 89.46 & 91.70 \\
$ATC_{N = 4}$ & 94.47 & 95.33 & 93.40 & 89.98 & 94.84 & 89.18 & 90.35 & 92.51 \\
\midrule
\multicolumn{9}{c}{Using CodeLlama13b as Detector LLM} \\
\midrule
Entropy & 41.51 & 41.56 & 53.72 & 43.46 & 59.20 & 54.63 & 44.87 & 48.42 \\
$log P(x)$ & 62.52 & 65.83 & 72.71 & 59.41 & 72.03 & 65.71 & 51.76 & 64.28 \\
LogRank & 59.84 & 62.61 & 69.11 & 56.88 & 66.09 & 60.56 & 45.83 & 60.13 \\
LRR & 46.07 & 45.37 & 46.53 & 43.81 & 36.23 & 36.13 & 27.19 & 40.19 \\
\midrule
$ATC_{N = 1}$ & 93.37 & 93.87 & 91.80 & 88.14 & 93.70 & 87.85 & 90.69 & 91.35 \\
$ATC_{N = 2}$ & 94.85 & 95.29 & 93.18 & 89.36 & 95.44 & 89.71 & 92.58 & 92.92 \\
$ATC_{N = 4}$ & \textbf{95.62} & \textbf{96.12} & \textbf{94.10} & \textbf{90.02} & \textbf{96.33} & \textbf{90.82} & \textbf{93.70} & \textbf{93.82} \\
\bottomrule
\end{tabular}
\end{table}

\subsection{Robustness to Comment Removal}
\label{section:removing_comments}
To test the robustness of our method in scenarios where comments and docstrings are unavailable, we evaluate detection performance after systematically removing them from the code. Table~\ref{tab:remove_comments} presents the results using CodeLlama-13B as the detector LLM.  
Our method remains highly effective, with minimal performance degradation, maintaining comparable results on MBPP and experiencing only a 3\% AUROC reduction on APPS, still outperforming previous methods with $N = 1$. Increasing $N$ further improves accuracy, reaching a mean AUROC of \textbf{94.16 on MBPP and 90.67 on APPS}. This demonstrates that our approach does not depend on comments, ensuring robustness against such transformations. %

\begin{table}[t]
\centering
\caption{Results when removing comments. Detector LLM is CodeLLama13b.}
\label{tab:remove_comments}
\begin{tabular}{lccccccccc}
\toprule
Generator & CLlama7b & CLlama13b & Gemma & Starchat & Claude & GPT3.5 & GPT4om & Avg. \\
\midrule
\multicolumn{9}{c}{MBPP} \\
\midrule
Entropy & 49.98 & 52.46 & 56.21 & 60.79 & 56.75 & 59.27 & 58.64 & 56.30 \\
$log P(x)$ & 68.03 & 70.73 & 67.67 & 72.05 & 71.92 & 73.94 & 72.88 & 71.03 \\
LogRank & 63.33 & 66.18 & 64.78 & 68.89 & 66.89 & 69.30 & 68.30 & 66.81 \\
LRR & 35.75 & 37.02 & 43.77 & 43.04 & 35.64 & 38.09 & 38.97 & 38.90 \\
\midrule
$ATC_{N = 1}$ & 93.92 & 94.68 & 86.29 & 90.59 & 95.09 & 92.76 & 94.51 & 92.55 \\
$ATC_{N = 2}$ & 94.82 & 95.50 & 87.69 & 91.65 & 95.88 & 94.01 & 95.18 & 93.53 \\
$ATC_{N = 4}$ & \textbf{95.54} & \textbf{95.94} & \textbf{88.50} & \textbf{92.34} & \textbf{96.44} & \textbf{94.48} & \textbf{95.85} & \textbf{94.16} \\
\midrule
\multicolumn{9}{c}{APPS} \\
\midrule
Entropy & 46.37 & 45.45 & 56.79 & 51.81 & 61.60 & 56.62 & 53.36 & 53.14 \\
$log P(x)$ & 59.31 & 60.48 & 70.09 & 60.58 & 72.50 & 67.58 & 56.69 & 63.89 \\
LogRank & 54.15 & 54.56 & 64.36 & 56.27 & 65.80 & 61.93 & 49.64 & 58.10 \\
LRR & 33.55 & 30.49 & 34.96 & 36.61 & 32.49 & 35.13 & 25.59 & 32.69 \\
\midrule
$ATC_{N = 1}$ & 87.24 & 88.37 & 88.08 & 83.57 & 92.78 & 86.07 & 88.70 & 87.83 \\
$ATC_{N = 2}$ & 89.04 & 90.31 & 89.60 & 84.74 & 94.36 & 87.96 & 90.65 & 89.53 \\
$ATC_{N = 4}$ & \textbf{90.68} & \textbf{91.28} & \textbf{90.72} & \textbf{85.73} & \textbf{95.28} & \textbf{89.08} & \textbf{91.89} & \textbf{90.67} \\
\bottomrule
\end{tabular}
\end{table}

\subsection{Evaluating the Approximated Task}  
\label{section:approximated_task}  
\begin{figure}
    \centering
    \begin{minipage}{0.25\textwidth}
        \centering
        \captionof{table}{Results using the original task with CodeLlama7b.}
        \label{tab:original_task}
        \begin{tabular}{lc}
            \toprule
            Method & Avg. \\
            \midrule
            \multicolumn{2}{c}{MBPP} \\
            \midrule
            Entropy & 55.76 \\
            $ATC_{N = 1}$ & 91.91 \\
            $ATC$ w/Task \;& 92.02 \\
            \midrule
            \multicolumn{2}{c}{APPS} \\
            \midrule
            Entropy & 51.82 \\
            $ATC_{N = 1}$ & 90.12 \\
            $ATC$ w/Task \;& 92.49 \\
            \bottomrule
        \end{tabular}
    \end{minipage}
    \hfill
    \begin{minipage}{0.72\textwidth}
        \centering
        \includegraphics[width=1\linewidth, trim=0cm 7.9cm 13.5cm 0cm, clip]{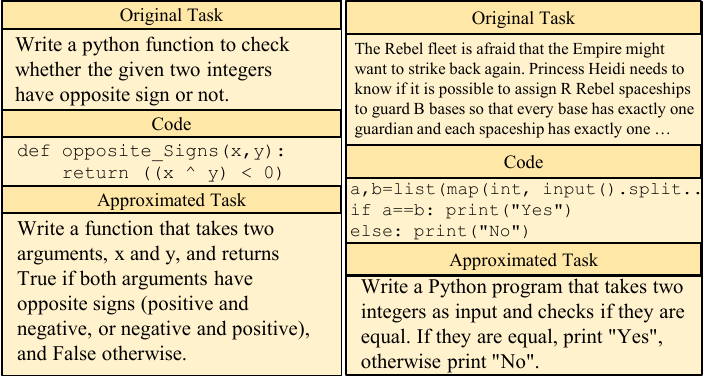}
        \caption{Approximated tasks examples. Left is MBPP, right is APPS. We present simple examples for readability.}
    \label{fig:task_examples}
    \end{minipage}
\end{figure}

We evaluate our approximated task by comparing detection performance against results obtained using the original task. As shown in Table~\ref{tab:original_task}, the mean AUROC across all \textit{generator LLMs} indicate a slight performance drop when using the approximated task instead of the original.
Figure~\ref{fig:task_examples} demonstrates that approximated tasks often contain slightly more detail than MBPP tasks. In contrast, for APPS, where tasks are longer and more descriptive, the approximated tasks tend to be more concise. These findings suggest that despite stylistic differences, the conditional distributions of the original and approximated tasks may still be similar. This is supported by Appendix D, which visualizes the conditional and unconditional probability distributions for the code snippets in Figure~\ref{fig:task_examples}.
Appendix E provides examples of approximated tasks from different seeds, showing how using $N>1$ averages slight variations in the conditional distribution. 

\begin{figure}[h]
    \centering
    \includegraphics[width=0.9\linewidth, trim=0.2cm 0.4cm 0.2cm 0.7cm, clip]{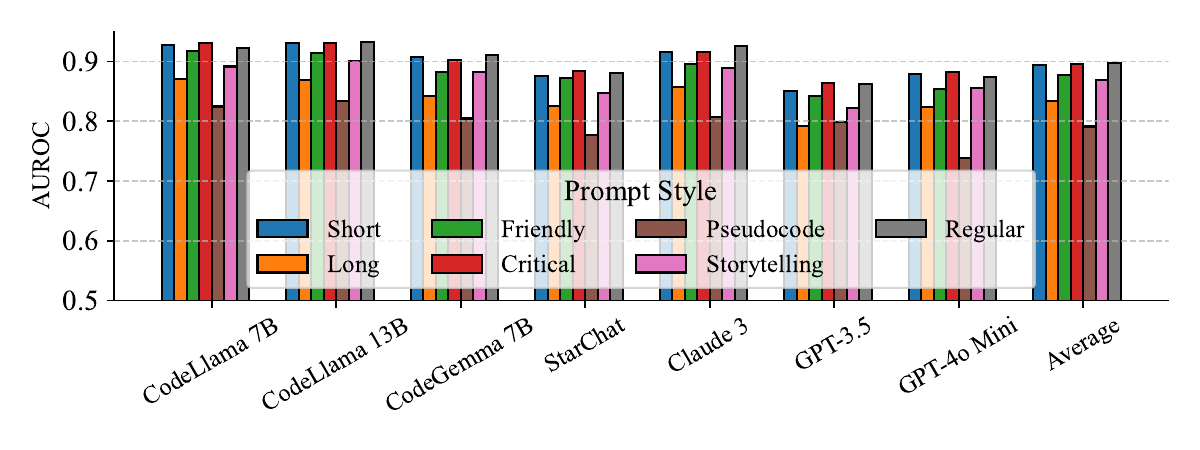}
    \caption{ATC with different prompting styles using CodeLlama7b.}
    \label{fig:prompt_styles}
\end{figure}

\subsection{Exploring Different Task Approximation Prompts}
\label{section:prompt_styles}

Here we explore the sensitivity of task approximation to different prompt styles. We aim to determine whether our method remains effective across various task approximation prompts or if performance is highly dependent on specific phrasing.
We test seven prompt styles, each offering a different way to approximate the task. \emph{Regular} provides a concise task description, while \emph{Short} enforces an even more minimal task. In contrast, \emph{Long} generates a verbose description. \emph{Storytelling} frames the task within a fictional scenario, \emph{Pseudocode} translates the code into a structured pseudocode, \emph{Friendly} offers a supportive tone, and \emph{Critical} delivers a specific and demanding specification.
In all experiments besides this one, we use the \emph{Regular} style.
Our experiment on APPS, selected for its notably descriptive tasks, reveals that prompts leading to \textbf{shorter and more accurate tasks} (\emph{Regular}, \emph{Short}, and \emph{Critical}) outperform those resulting in longer tasks (\emph{Long}, \emph{Pseudocode}). We observed that \emph{Storytelling} occasionally produced vague or incorrect tasks, likely due to the limitations of the relatively small detector LLM.
Results are in Figure~\ref{fig:prompt_styles}. Full prompts and examples are in Appendix F.

\subsection{Effects of Decoding Strategies}
\begin{figure}[h]
    \centering
    \begin{minipage}{0.38\textwidth}
        \centering
        \includegraphics[width=0.66\linewidth, trim=0.35cm 0.38cm 0.42cm 1cm, clip]{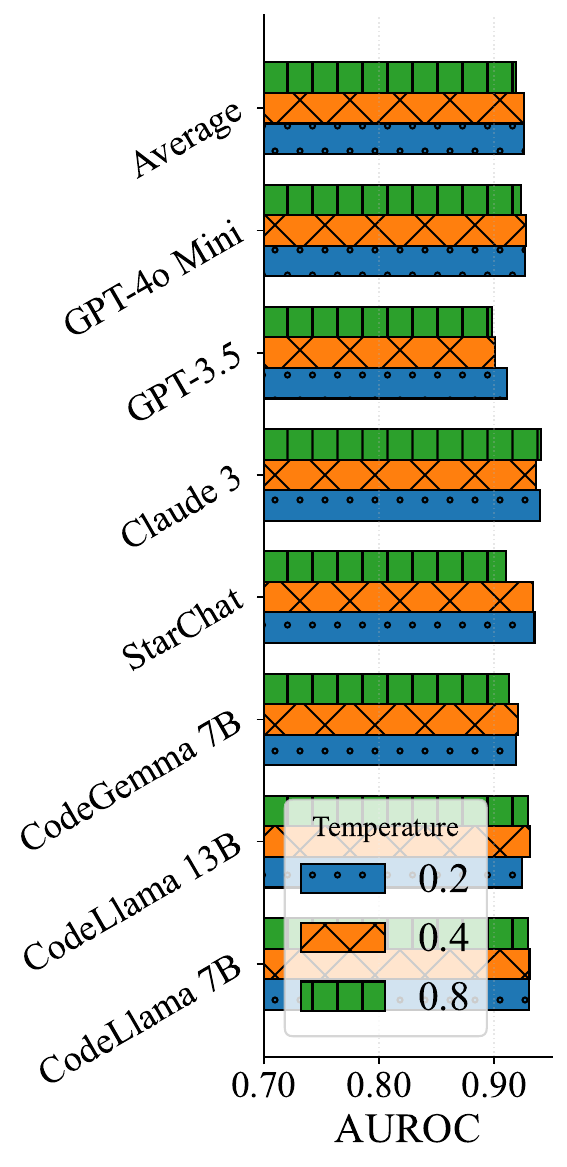}
        \caption{Temperature effects on MBPP with CodeLlama7b.}
        \label{fig:temperatures}
    \end{minipage}
    \hfill
    \begin{minipage}{0.61\textwidth}
        \centering
        \captionof{table}{CodeContest results with CodeLlama7b.}
        \begin{tabular}{lcccc}
        \toprule
        Lang. & Method & CLlama13b & Starchat & GPT3.5 \\
        \midrule
        \multirow{9}{*}{CPP} & Shi \cite{shi2025detectcodegpt} & 81.07 & 73.59 & 81.96 \\
        & Ye \cite{ye2024uncovering} & 89.87 & 83.42 & 90.82 \\
        \cmidrule{2-5}
        & Entropy & 29.83 & 39.20 & 43.29 \\
        & $log P(x)$ & 68.80 & 65.23 & 72.11 \\
        & LogRank & 67.33 & 65.35 & 71.37 \\
        & LRR~\cite{su2023detectllm} & 51.09 & 56.51 & 57.68 \\
        \cmidrule{2-5}
        & $ATC_{N = 1}$ & 97.63 & 90.94 & 92.99 \\
        & $ATC_{N = 2}$ & 98.26 & 91.52 & 93.97 \\
        & $ATC_{N = 4}$ & \textbf{98.44} & \textbf{91.82} & \textbf{94.69} \\
        \midrule
        \multirow{5}{*}{Java} & Shi \cite{shi2025detectcodegpt} & 76.65 & 70.72 & 82.10 \\
        & Yang \cite{yang2023zero} & - & - & 64.03 \\
        \cmidrule{2-5}
        & $ATC_{N = 1}$ & 92.30 & 89.61 & 91.73 \\
        & $ATC_{N = 2}$ & 92.54 & 90.48 & 93.02 \\
        & $ATC_{N = 4}$ & \textbf{92.93} & \textbf{90.85} & \textbf{93.00} \\
        \bottomrule
        \end{tabular}
        \label{tab:cpp_java}
    \end{minipage}
\end{figure}

We evaluate the robustness of \textit{ATC} by examining the impact of decoding temperatures on detection performance, using the same configuration as in~\cite{ye2024uncovering}. Higher temperatures introduce greater variability in the generated outputs, while lower temperatures yield more deterministic results (See Appendix B). Although entropy-based scoring methods might be sensitive to varying temperatures, the results in Figure~\ref{fig:temperatures} show that our performance remains robust across a range of values. Nonetheless, we do observe a slight decline in mean AUROC at higher temperatures, suggesting that high variability can impact detection accuracy.

\subsection{Generalization to Other Programming Languages}
\label{section:cpp_java}

To assess the generalization of \textit{ATC} across programming languages, which is critical for real-world applications, we experiment on CPP and Java using the CodeContest dataset~\cite{li2022competition}. We identify 152 CPP instances and 129 Java instances in the test set. Results in Table~\ref{tab:cpp_java} focus on \textit{generator LLMs} from previous works due to space constraints, with full results and comparisons in Appendix G. Our method consistently outperforms other approaches across all \textit{generator LLMs}, demonstrating its ability to generalize across different programming languages.

\subsection{Impact of Code Length}
We examine how code length affects detection performance using APPS, where solutions are generally longer than those in MBPP. We measure length in terms of the number of characters and group each sample—whether human-written or LLM-generated—into its corresponding length interval, independent of the original task. Consistent with previous findings, our results in Figure~\ref{fig:code_length} using CodeLlama7b show that detection performance improves as code length increases, likely due to greater certainty in token predictions as the code progresses. This suggests that in practical real-world scenarios, where code is typically longer, our method is expected to perform well.

\subsection{Real-World Considerations}
\label{section:real_world}
\begin{figure}[t]
    \centering
    \begin{minipage}{0.51\textwidth}
        \centering
        \captionof{figure}{Number of generated tokens in different methods.}
        \label{fig:num_gen_tokens}
        \centering
        \includegraphics[width=1\linewidth, trim=0.2cm 0.4cm 0cm 0.4cm, clip]{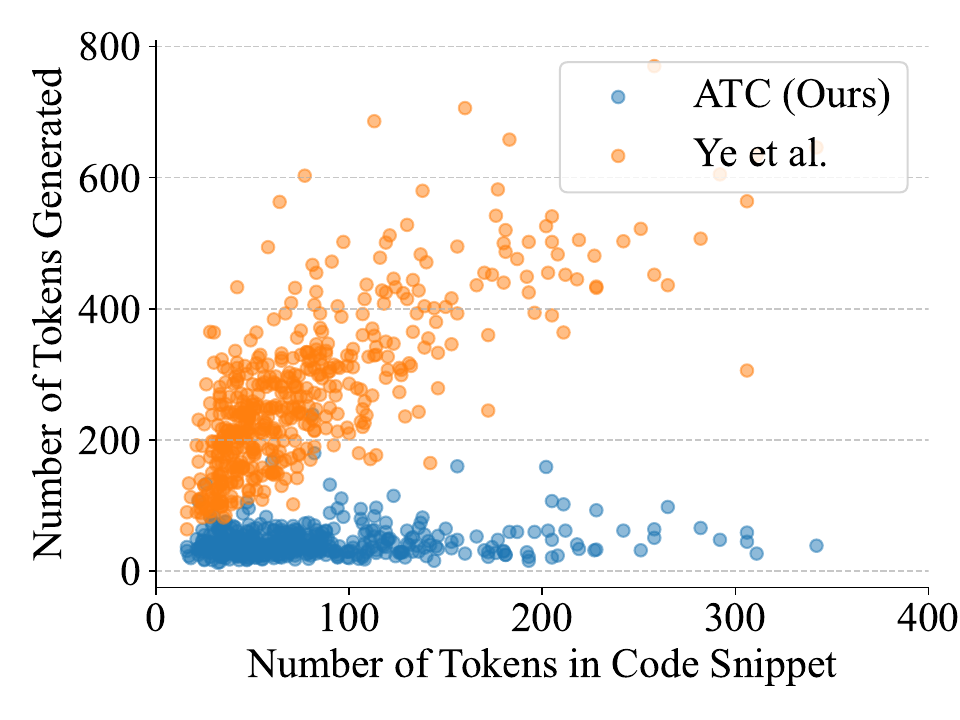}
    \end{minipage}
    \hfill
    \begin{minipage}{0.46\textwidth}
        \centering
        \captionof{table}{Recall @ FPR Results.}
        \label{tab:recall_at_fpr}
 \begin{tabular}{lcc}
        \toprule
        Method & \multicolumn{2}{c}{Recall @ FPR 10\% Avg.} \\
        \cmidrule(lr){2-3}
               & MBPP & APPS \\
        \midrule
        \multicolumn{3}{c}{CodeLlama13b as Detector LLM} \\
        \midrule
        Entropy & 10.09 & 9.96 \\
        $log P(x)$ & 31.23 & 24.58 \\
        LogRank & 30.16 & 21.28 \\
        LRR~\cite{su2023detectllm} & 19.18 & 7.01 \\
        \midrule
        $ATC_{N=1}$  & 76.12 & 77.72 \\
        $ATC_{N=2}$  & 80.37 & 81.58 \\
        $ATC_{N=4}$  & \textbf{83.92} & \textbf{84.08} \\
        \bottomrule
    \end{tabular}
    \end{minipage}
\end{figure}

\noindent\textbf{Limitations of AUROC} 
In practical settings, detection accuracy measured by AUROC may not fully reflect operational efficacy. Here we analyze our method's recall (true positive rate) at a fixed false positive rate (FPR). This evaluation better captures the trade-offs relevant to real-world production scenarios, ensuring reliable identification of LLM-generated code while minimizing false alarms on human-written code. As shown in Table~\ref{tab:recall_at_fpr}, our method achieves a recall of roughly 84\% at a false positive rate of 10\%, demonstrating strong detection capability with minimal misclassifications. Full results are in Appendix H.

\bigskip
\noindent\textbf{Analyzing the Number of Generated Tokens}
We compare the complexity characteristics of our method with the previous SOTA \cite{ye2024uncovering}. While both approaches rely on querying an LLM multiple times per sample, they differ significantly in the nature and length of the generated outputs. In both cases, the primary latency bottleneck lies in the generation step itself. \cite{ye2024uncovering} prompts the LLM with \textit{"Please explain the functionality of the given code, then rewrite it in a single markdown code block."}. This yields a combined output containing a detailed natural language explanation followed by a full code rewrite. 
In contrast, our method only requires a concise task approximation. Notably, the length of this generated task remains roughly constant regardless of the input code size, compared to \cite{ye2024uncovering}, where the output length scales linearly with the input. To quantify this, we measure the number of generated tokens on MBPP (see Figure~\ref{fig:num_gen_tokens}). As a result, our method is not only faster but also more cost-efficient in settings that rely on third-party APIs where pricing is based on the number of generated tokens. For example, the average generation time using CodeLlama-7b per MBPP sample is \textbf{0.99} seconds in ATC, compared to \textbf{6.04} seconds in the other approach. Latency was measured on a single Nvidia RTX6000 GPU.

\subsection{Ablation Experiments}

\begin{figure}[t]
    \centering
    \begin{minipage}{0.48\textwidth}
        \centering
        \includegraphics[width=1\linewidth, trim=0.2cm 0.4cm 0.2cm 0.4cm, clip]{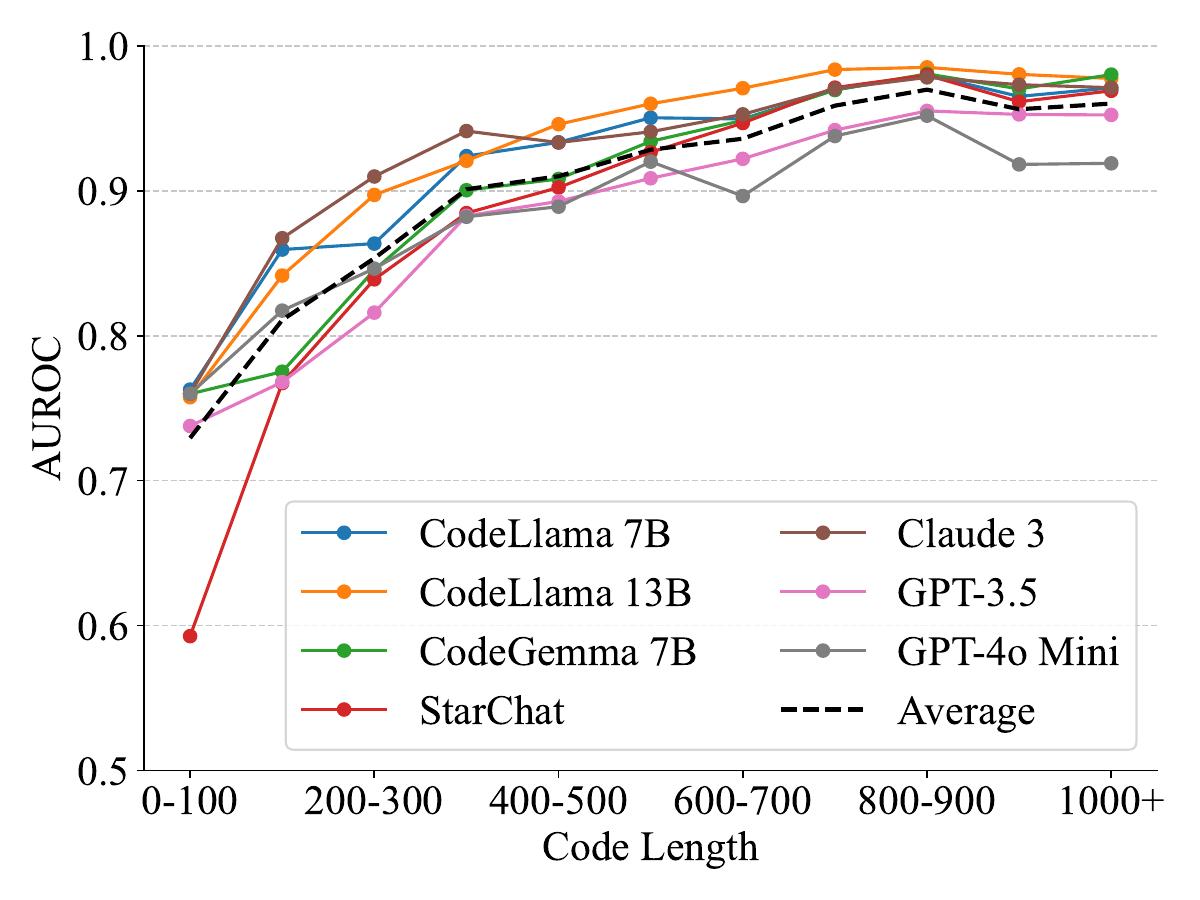}
        \caption{Impact of code length on APPS.}
        \label{fig:code_length}
    \end{minipage}
    \hfill
    \begin{minipage}{0.50\textwidth}
        \centering
        \includegraphics[width=0.98\linewidth, trim=0.2cm 0.4cm 0.2cm 0.4cm, clip]{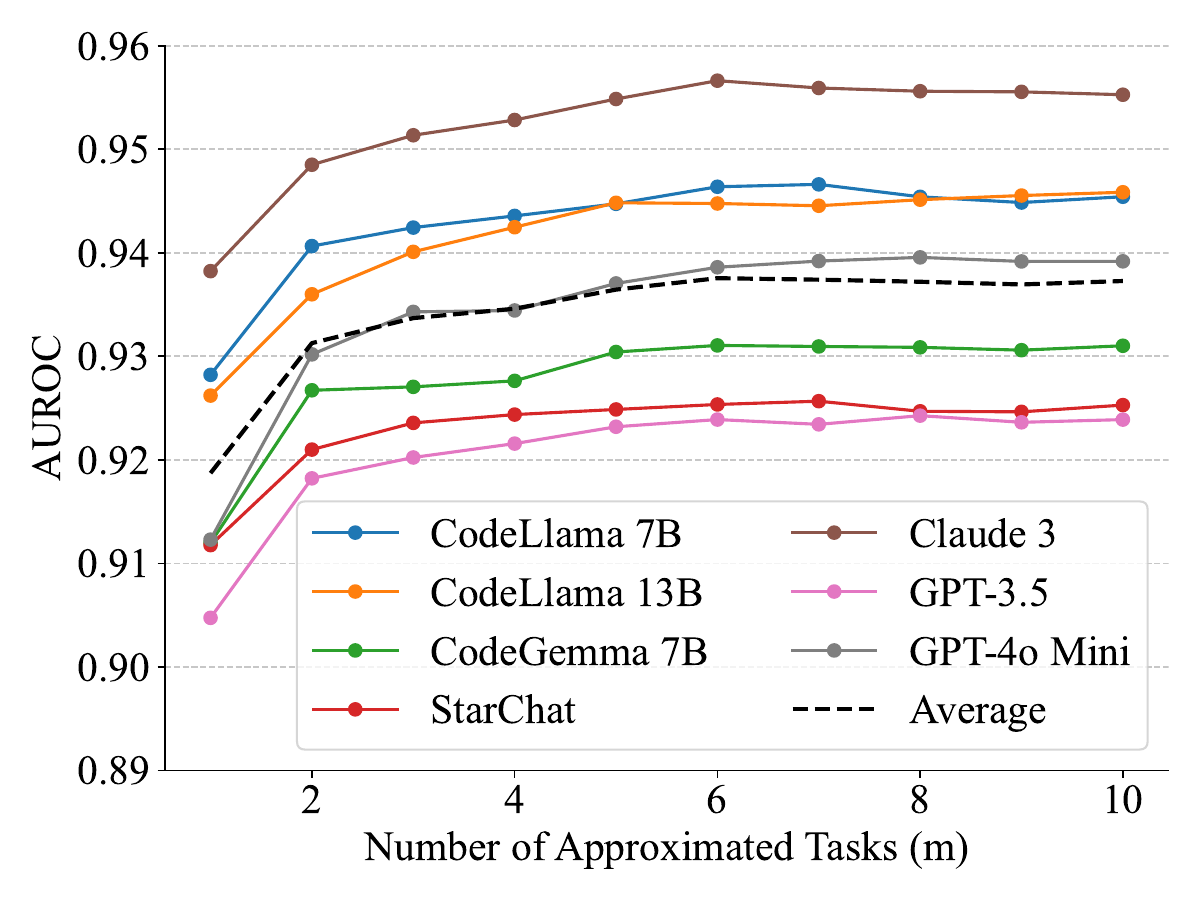}
        \caption{Effects of increasing $N$ on MBPP.}
        \label{fig:increasing_n}
    \end{minipage}
\end{figure}

\noindent\textbf{Increasing the Number of Task Approximations}
In most experiments we use $N \le 4$, however increasing $N$ further enhances results.
Figure~\ref{fig:increasing_n}, using CodeLlama7b, demonstrates that performance gains scale with the number of tasks, with the most significant improvement occurring at $N = 2$ and diminishing returns appearing from $N = 4$. 
The choice of $N$ presents a tradeoff between accuracy and runtime, and should be adjusted based on real-world constraints.

\bigskip
\noindent\textbf{Comparison with Alternative Scoring Methods}
\label{ablation:scoring_methods} We replace entropy with alternative scoring methods—mean $log P(x)$, LogRank, and LRR—while keeping the task approximation framework unchanged.
Table~\ref{tab:alternative_scores} presents average AUROC across all \textit{generator LLMs}, showing that entropy is the most effective scoring method. Entropy captures global uncertainty over the full output distribution, while alternative methods rely on token-level likelihoods or ranks, making them more sensitive to local variations.
Among these, LogRank performs best, indicating that ranks may provide a stronger signal than raw likelihoods. However, entropy remains the overall best scoring method across our experiments.

\bigskip
\noindent\textbf{Score Computation with Comment Tokens}
We conduct an ablation study where we include comments in the token entropy calculation instead of excluding them. As shown in Table~\ref{tab:including_comments}, this results in a consistent drop in average AUROC across all \textit{generator LLMs}. This degradation aligns with our hypothesis that comments often serve as implicit task descriptions within the code snippet. Including them in entropy computation disrupts the separation between task conditioning and code token certainty, weakening detection results.

\begin{figure}[t]
    \centering
    \begin{minipage}{0.49\textwidth}
        \centering
        \captionof{table}{Results with alternative scoring methods.}
        \label{tab:alternative_scores}
        \begin{tabular}{lcc}
        \toprule
        Method & \multicolumn{2}{c}{AUROC Avg.} \\
        \cmidrule(lr){2-3}
         & MBPP & APPS \\
        \midrule
        \multicolumn{3}{c}{CodeLlama7b as Detector LLM} \\
        \midrule
        $ATC_{N=1}$ w/$log P(x)$ & 80.09\textsubscript{\textcolor{black}{(-11.82)}} & 84.14\textsubscript{\textcolor{black}{(-5.98)}} \\
        $ATC_{N=1}$ w/LogRank & 87.08\textsubscript{\textcolor{black}{(-4.83)}} & 76.00\textsubscript{\textcolor{black}{(-14.12)}} \\
        $ATC_{N=1}$ w/LRR & 81.38\textsubscript{\textcolor{black}{(-10.53)}} & 80.45\textsubscript{\textcolor{black}{(-9.67)}} \\
        \midrule
        \multicolumn{3}{c}{CodeLlama13b as Detector LLM} \\
        \midrule
        $ATC_{N=1}$ w/$log P(x)$ & 83.94\textsubscript{\textcolor{black}{(-8.17)}} & 80.78\textsubscript{\textcolor{black}{(-10.57)}} \\
        $ATC_{N=1}$ w/LogRank & 89.08\textsubscript{\textcolor{black}{(-3.03)}} & 86.06\textsubscript{\textcolor{black}{(-5.29)}} \\
        $ATC_{N=1}$ w/LRR & 82.51\textsubscript{\textcolor{black}{(-9.60)}} & 83.07\textsubscript{\textcolor{black}{(-8.28)}} \\
        \bottomrule
        \end{tabular}
    \end{minipage}
    \hfill
    \begin{minipage}{0.46\textwidth}
        \centering
        \captionof{table}{Results when including comment tokens in score calculation.}
        \label{tab:including_comments}
        \begin{tabular}{lcc}
        \toprule
        Method & \multicolumn{2}{c}{AUROC Avg.} \\
        \cmidrule(lr){2-3}
         & MBPP & APPS \\
        \midrule
        \multicolumn{3}{c}{CodeLlama7b as Detector LLM} \\
        \midrule
        $ATC_{N = 1}$ & 90.10\textsubscript{\textcolor{black}{(-1.81)}} & 87.24\textsubscript{\textcolor{black}{(-2.88)}} \\
        $ATC_{N = 2}$ & 91.46\textsubscript{\textcolor{black}{(-1.7)}} & 88.97\textsubscript{\textcolor{black}{(-2.73)}} \\
        $ATC_{N = 4}$ & 91.91\textsubscript{\textcolor{black}{(-1.62)}} & 89.90\textsubscript{\textcolor{black}{(-2.61)}} \\
        \midrule
        \multicolumn{3}{c}{CodeLlama13b as Detector LLM} \\
        \midrule
        $ATC_{N = 1}$ & 90.63\textsubscript{\textcolor{black}{(-1.48)}} & 88.81\textsubscript{\textcolor{black}{(-2.54)}} \\
        $ATC_{N = 2}$ & 92.01\textsubscript{\textcolor{black}{(-1.39)}} & 90.57\textsubscript{\textcolor{black}{(-2.35)}} \\
        $ATC_{N = 4}$ & 92.94\textsubscript{\textcolor{black}{(-1.28)}} & 91.64\textsubscript{\textcolor{black}{(-2.18)}} \\
        \bottomrule
        \end{tabular}
    \end{minipage}
\end{figure}

\section{Conclusion and Future Work}
As LLMs become increasingly prevalent in coding tasks, their associated social and ethical risks demand reliable detection methods. We identify a key challenge: distinguishing human-written from LLM-generated code is fundamentally different from natural text when relying solely on the unconditional probability distribution. 
To address this, we introduce a novel, simple, and zero-shot approach that approximates the conditional probability distribution using task approximation, followed by an entropy-based scoring algorithm. Our method outperforms previous approaches across all relevant benchmarks and demonstrates robustness through extensive experiments and ablation studies. Furthermore, its simplicity enables future integration with other probability-based detection methods.
Our analysis is currently limited to publicly available benchmarks; however, while our results are promising, task approximation in domain-specific repositories may pose additional challenges and warrants further study. 
In future work, we plan to extend our approach to detect edited LLM-generated code and explore robustness against adversarial attacks. Finally, an additional promising direction is to improve the quality of task approximations, potentially through reflective reasoning capabilities.

\begin{credits}

\subsubsection{\ackname} 
The authors thank the Israeli Council for Higher Education (CHE) via the Data Science Research Center and the Lynn and William Frankel Center for Computer Science at BGU. 

\subsubsection{\discintname}
The authors have no competing interests to declare that are
relevant to the content of this article. 
\end{credits}
\newpage

\bibliographystyle{splncs04}
\bibliography{bibliography}

\begin{thebibliography}{10}
\providecommand{\url}[1]{\texttt{#1}}
\providecommand{\urlprefix}{URL }
\providecommand{\doi}[1]{https://doi.org/#1}

\bibitem{anthropic2024}
Anthropic (2024), \url{https://www.anthropic.com/}

\bibitem{openai2024}
Openai (2024), \url{https://openai.com/api}

\bibitem{stackoverflow2024survey}
Stackoverflow developer survey (2024), \url{https://survey.stackoverflow.co/2024/}

\bibitem{meta_llama3.1}
Meta ai, llama 3.1 (2024), \url{https://llama.meta.com/}

\bibitem{austin2021program}
Austin, J., Odena, A., Nye, M., Bosma, M., Michalewski, H., Dohan, D., Jiang, E., Cai, C., Terry, M., Le, Q., et~al.: Program synthesis with large language models. arXiv preprint arXiv:2108.07732  (2021)

\bibitem{bakhtin2019real}
Bakhtin, A., Gross, S., Ott, M., Deng, Y., Ranzato, M., Szlam, A.: Real or fake? learning to discriminate machine from human generated text. arXiv preprint arXiv:1906.03351  (2019)

\bibitem{bao2023fast}
Bao, G., Zhao, Y., Teng, Z., Yang, L., Zhang, Y.: Fast-detectgpt: Efficient zero-shot detection of machine-generated text via conditional probability curvature. arXiv preprint arXiv:2310.05130  (2023)

\bibitem{bommasani2021opportunities}
Bommasani, R., Hudson, D.A., Adeli, E., Altman, R., Arora, S., von Arx, S., Bernstein, M.S., Bohg, J., Bosselut, A., Brunskill, E., et~al.: On the opportunities and risks of foundation models. arXiv preprint arXiv:2108.07258  (2021)

\bibitem{demirok2024aigcodeset}
Demirok, B., Kutlu, M.: Aigcodeset: A new annotated dataset for ai generated code detection. arXiv preprint arXiv:2412.16594  (2024)

\bibitem{gehrmann2019gltr}
Gehrmann, S., Strobelt, H., Rush, A.: Gltr: Statistical detection and visualization of generated text. In: Proceedings of the 57th Annual Meeting of the Association for Computational Linguistics: System Demonstrations. Association for Computational Linguistics (2019)

\bibitem{hendrycks2021measuring}
Hendrycks, D., Basart, S., Kadavath, S., Mazeika, M., Arora, A., Guo, E., Burns, C., Puranik, S., He, H., Song, D., Steinhardt, J.: Measuring coding challenge competence with {APPS}. In: Thirty-fifth Conference on Neural Information Processing Systems Datasets and Benchmarks Track (Round 2) (2021)

\bibitem{holtzmancurious}
Holtzman, A., Buys, J., Du, L., Forbes, M., Choi, Y.: The curious case of neural text degeneration. In: International Conference on Learning Representations

\bibitem{huang2024gpt}
Huang, X.Y., Vishnubhotla, K., Rudzicz, F.: The gpt-writingprompts dataset: A comparative analysis of character portrayal in short stories. arXiv preprint arXiv:2406.16767  (2024)

\bibitem{jawahar2020automatic}
Jawahar, G., Abdul-Mageed, M., Laks~Lakshmanan, V.: Automatic detection of machine generated text: A critical survey. In: Proceedings of the 28th International Conference on Computational Linguistics. pp. 2296--2309 (2020)

\bibitem{lee2024wrote}
Lee, T., Hong, S., Ahn, J., Hong, I., Lee, H., Yun, S., Shin, J., Kim, G.: Who wrote this code? watermarking for code generation. In: Proceedings of the 62nd Annual Meeting of the Association for Computational Linguistics (Volume 1: Long Papers). pp. 4890--4911 (2024)

\bibitem{li2022competition}
Li, Y., Choi, D., Chung, J., Kushman, N., Schrittwieser, J., Leblond, R., Eccles, T., Keeling, J., Gimeno, F., Dal~Lago, A., et~al.: Competition-level code generation with alphacode. Science  \textbf{378}(6624),  1092--1097 (2022)

\bibitem{mitchell2023detectgpt}
Mitchell, E., Lee, Y., Khazatsky, A., Manning, C.D., Finn, C.: Detectgpt: Zero-shot machine-generated text detection using probability curvature. In: International Conference on Machine Learning. pp. 24950--24962. PMLR (2023)

\bibitem{mitrovic2023chatgpt}
Mitrovi{\'c}, S., Andreoletti, D., Ayoub, O.: Chatgpt or human? detect and explain. explaining decisions of machine learning model for detecting short chatgpt-generated text. arXiv preprint arXiv:2301.13852  (2023)

\bibitem{pu2023deepfake}
Pu, J., Sarwar, Z., Abdullah, S.M., Rehman, A., Kim, Y., Bhattacharya, P., Javed, M., Viswanath, B.: Deepfake text detection: Limitations and opportunities. In: 2023 IEEE Symposium on Security and Privacy (SP). pp. 1613--1630 (2023)

\bibitem{raffel2020exploring}
Raffel, C., Shazeer, N., Roberts, A., Lee, K., Narang, S., Matena, M., Zhou, Y., Li, W., Liu, P.J.: Exploring the limits of transfer learning with a unified text-to-text transformer. Journal of machine learning research  \textbf{21}(140),  1--67 (2020)

\bibitem{roziere2023code}
Roziere, B., Gehring, J., Gloeckle, F., Sootla, S., Gat, I., Tan, X.E., Adi, Y., Liu, J., Sauvestre, R., Remez, T., et~al.: Code llama: Open foundation models for code. arXiv preprint arXiv:2308.12950  (2023)

\bibitem{shi2025detectcodegpt}
Shi, Y., Zhang, H., Wan, C., Gu, X.: Between lines of code: Unraveling the distinct patterns of machine and human programmers. In: Proceedings of the 47th International Conference on Software Engineering (ICSE 2025). IEEE (2025)

\bibitem{su2023detectllm}
Su, J., Zhuo, T.Y., Wang, D., Nakov, P.: Detectllm: Leveraging log rank information for zero-shot detection of machine-generated text. arXiv preprint arXiv:2306.05540  (2023)

\bibitem{team2024codegemma}
Team, C., Zhao, H., Hui, J., Howland, J., Nguyen, N., Zuo, S., Hu, A., Choquette-Choo, C.A., Shen, J., Kelley, J., et~al.: Codegemma: Open code models based on gemma. arXiv preprint arXiv:2406.11409  (2024)

\bibitem{thomas2008detecting}
Thomas, L.: Detecting fack content with relative entropy scoring. In: CEUR Workshop Proceedings, ECAI'08 Workshop on Plagiarism Analysis, Authorship Identification and Near-Duplication Detection, November. vol.~377, pp. 27--31 (2008)

\bibitem{Tunstall2023starchat-alpha}
Tunstall, L., Lambert, N., Rajani, N., Beeching, E., Le~Scao, T., von Werra, L., Han, S., Schmid, P., Rush, A.: Creating a coding assistant with starcoder. Hugging Face Blog  (2023), https://huggingface.co/blog/starchat-alpha

\bibitem{vaswani2017attention}
Vaswani, A., Shazeer, N., Parmar, N., Uszkoreit, J., Jones, L., Gomez, A.N., Kaiser, {\L}., Polosukhin, I.: Attention is all you need. Advances in neural information processing systems  \textbf{30} (2017)

\bibitem{wang-etal-2021-codet5}
Wang, Y., Wang, W., Joty, S., Hoi, S.C.: {CodeT5}: Identifier-aware unified pre-trained encoder-decoder models for code understanding and generation. Proceedings of the 2021 Conference on Empirical Methods in Natural Language Processing  (2021)

\bibitem{xu2024investigating}
Xu, J., Zhang, H., Yang, Y., Cheng, Z., Lyu, J., Liu, B., Zhou, X., Yang, L., Bacchelli, A., Chiam, Y.K., et~al.: Investigating efficacy of perplexity in detecting llm-generated code. arXiv preprint arXiv:2412.16525  (2024)

\bibitem{xu2024detecting}
Xu, Z., Sheng, V.S.: Detecting ai-generated code assignments using perplexity of large language models. Proceedings of the aaai conference on artificial intelligence  \textbf{38}(21),  23155--23162 (2024)

\bibitem{yangdna}
Yang, X., Cheng, W., Wu, Y., Petzold, L., Wang, W.Y., Chen, H.: {DNA-GPT}: Divergent n-gram analysis for training-free detection of gpt-generated text. The Twelfth International Conference on Learning Representations (ICLR)  (2024)

\bibitem{yang2023zero}
Yang, X., Zhang, K., Chen, H., Petzold, L., Wang, W.Y., Cheng, W.: Zero-shot detection of machine-generated codes. arXiv preprint arXiv:2310.05103  (2023)

\bibitem{ye2024uncovering}
Ye, T., Du, Y., Ma, T., Wu, L., Zhang, X., Ji, S., Wang, W.: Uncovering llm-generated code: A zero-shot synthetic code detector via code rewriting. arXiv preprint arXiv:2405.16133  (2024)

\bibitem{yu2023gpt}
Yu, X., Qi, Y., Chen, K., Chen, G., Yang, X., Zhu, P., Zhang, W., Yu, N.: Gpt paternity test: Gpt generated text detection with gpt genetic inheritance. CoRR  (2023)

\bibitem{zellers2019defending}
Zellers, R., Holtzman, A., Rashkin, H., Bisk, Y., Farhadi, A., Roesner, F., Choi, Y.: Defending against neural fake news. Advances in neural information processing systems  \textbf{32} (2019)

\bibitem{zhong2020neural}
Zhong, W., Tang, D., Xu, Z., Wang, R., Duan, N., Zhou, M., Wang, J., Yin, J.: Neural deepfake detection with factual structure of text. In: Conference on Empirical Methods in Natural Language Processing (EMNLP) (2020)

\end{thebibliography}

\appendix

\title{Appendix}
\author{}
\institute{}
\maketitle

\section{Initial Experiment Setup and Methodology}  
We provide additional details on the experimental setup described in Section 1.  

\subsection{Datasets}  
We use two datasets to compare LLM-generated and human-written content:  
\begin{itemize}  
    \item \textbf{MBPP (Mostly Basic Python Problems)~\cite{austin2021program}}: A collection of 1000 entry-level Python programming tasks with human-written solutions. We use the test set (500 tasks), treating problem descriptions as task prompts and the corresponding solutions as human-written code samples.  
    \item \textbf{WritingPrompts~\cite{huang2024gpt}}: A dataset of approximately 300k creative writing responses to prompts from Reddit\footnote{https://www.reddit.com/r/WritingPrompts/?rdt=48490}. We treat the prompts as task descriptions and the corresponding responses as human-written text. To ensure a balanced comparison with MBPP, we sample 500 prompts and responses from the test set.  
\end{itemize}   

\subsection{LLM-Generated Responses}  
We generate LLM-written samples using:  
\begin{itemize}  
    \item \textbf{CodeLlama-7B}~\cite{roziere2023code} for MBPP, generating code completions based on the provided task descriptions, using the chain-of-thought (COT) code generation prompt from~\cite{ye2024uncovering}.  
    \item \textbf{LLaMA 3.1-8B}~\cite{meta_llama3.1} for WritingPrompts, generating natural language responses using the prompt:  
    \textit{"You serve as a writing assistant. I will first give you a prompt. You need to tell me a story about the prompt. <PROMPT>"}  
\end{itemize}  

While we present results using these models, we observe similar trends when using LLaMA 3.1 as the generator for MBPP tasks.
To control for response length, we discard responses shorter than 200 characters and truncate longer responses to 200 characters.  
All model generations use a temperature of 0.7 and $top_p$ sampling of 0.95.  

\subsection{Entropy Computation}  
For each sample, we compute \textbf{mean token entropy} under two settings:

\textbf{Unconditional Setting} (no task information)
\[
H_{\text{uncond}} = -\frac{1}{m} \sum_{k=1}^{m} \sum_{v \in \mathcal{V}} P(v \mid x_{<k}) \log P(v \mid x_{<k})
\]

\textbf{Task-Conditioned Setting} (given the original task prompt) 
\[
H_{\text{cond}} = -\frac{1}{m} \sum_{k=1}^{m} \sum_{v \in \mathcal{V}} P(v \mid x_{<k}, t_i) \log P(v \mid x_{<k}, t_i)
\]

where \( m \) is the total number of tokens in the generated sequence, \( \mathcal{V} \) is the vocabulary of the language model, and \( P(v \mid x_{<k}, t_i) \) represents the probability assigned by the model to token \( v \) at position \( k \), given the preceding tokens \( x_{<k} \) and, optionally, the task prompt \( t_i \).  
Mean token entropy is computed using the same model that generated each sample, allowing us to analyze the impact of task conditioning on entropy differences.  

\newpage
\section{Sampling Parameters: $top_p$ and Temperature}
\label{appendix:sampling}

When sampling from a generative language model, two of the main parameters that control the diversity and randomness of generated text are $top_p$ (nucleus sampling) and temperature.

$top_p$ (nucleus) sampling~\cite{holtzmancurious} dynamically limits the number of tokens considered in the sampling step by selecting the smallest subset of tokens, whose cumulative probability exceeds the predefined threshold $p$.

The temperature, denoted by $T$, controls the randomness of token selection by adjusting the probability distribution before sampling. Given the original logits $z_i$ produced by the model, the token probabilities are computed using the softmax function:

\[
    P(x_i) = \frac{\exp(z_i / T)}{\sum_j \exp(z_j / T)}
\]

Lower temperatures make the distribution sharper, focusing on high-probability tokens, thus making the output more deterministic. Higher temperatures flatten the distribution, increasing the likelihood of sampling lower-probability tokens, thus adding more randomness.

\newpage
\section{Prevalence of Comments in Code}

\begin{figure}[h]
    \centering
    \includegraphics[width=0.98\linewidth, trim=0.2cm 0.4cm 0.2cm 0.8cm, clip]{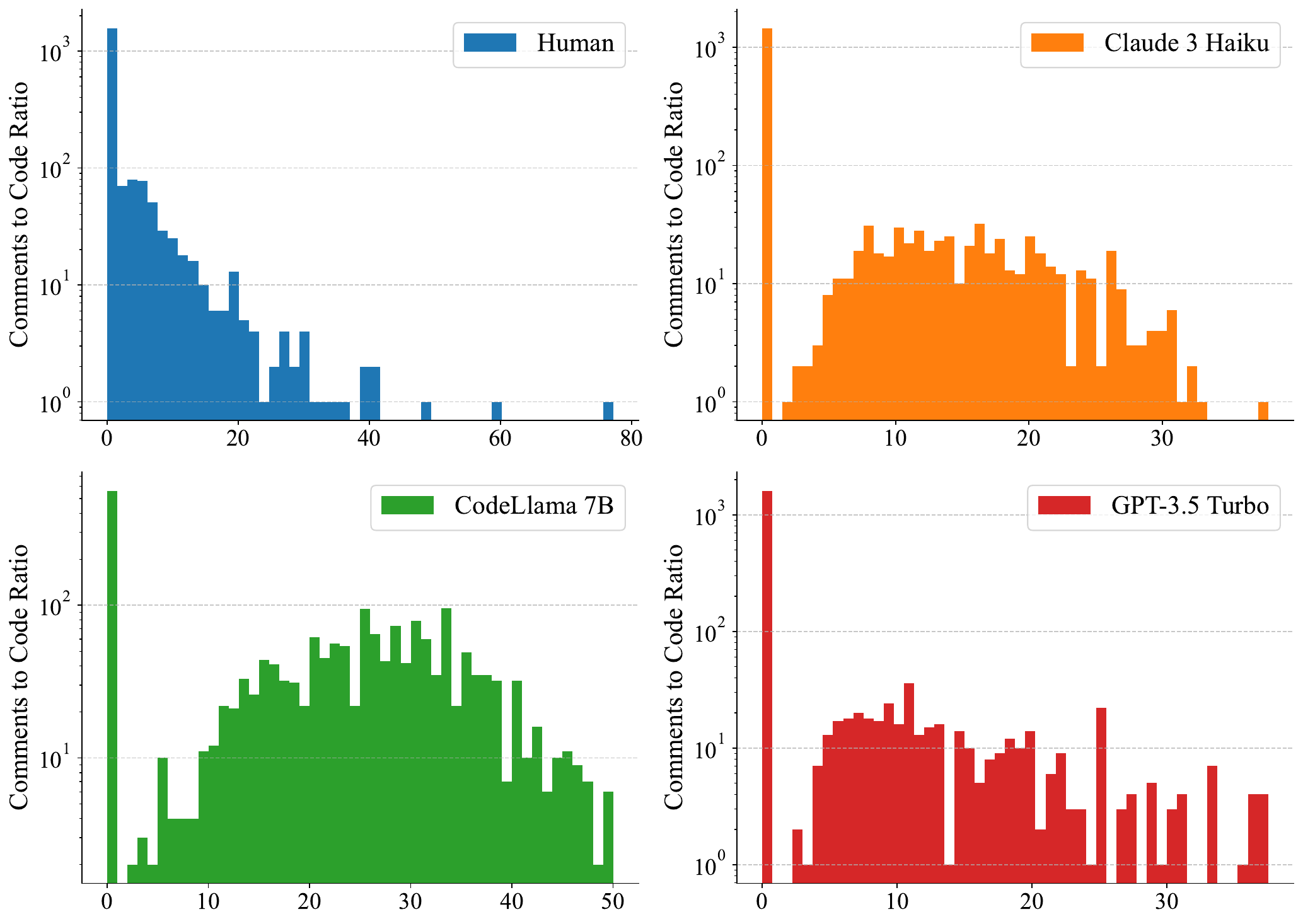}
\caption{Distribution of the comments-to-code ratio across human-written and LLM-generated code samples. Each subplot represents a different source: Human-written code (top-left), Claude 3 Haiku (top-right), CodeLlama 7B (bottom-left), and GPT-3.5 Turbo (bottom-right). The y-axis is plotted on a logarithmic scale for better visualization of the distribution. }
\end{figure}

We analyze the prevalence of comments in code by comparing human-written and LLM-generated solutions in the APPS dataset. Specifically, we compute the log ratio of comment lines to code lines in each sample.  

Our analysis reveals that both humans and LLMs frequently use comments, underscoring their integral role in code. This motivates our careful treatment of comments throughout the paper, mainly by leveraging them for additional task conditioning, but also evaluating our method's robustness when removing them entirely as a preprocessing step.

\newpage
\section{Visualizing Probability Distributions}
We visualize token probability distributions for two code snippets, one from MBPP and one from APPS, using CodeLlama-7B as the detector LLM. We consider three settings:

\begin{enumerate}
\item \textbf{Conditional} probability distribution given the \textbf{original task}.
\item \textbf{Conditional} probability distribution given the \textbf{approximated task}.
\item \textbf{Unconditional} probability distribution (no task conditioning).
\end{enumerate}

\subsection{MBPP Example}
As discussed in Section 4.4, MBPP tasks are short and simple, resulting in approximated tasks that closely resemble the originals. Unsurprisingly, the conditional probability distributions for the approximated and original tasks are similar, while the unconditional distribution is significantly different.

\bigskip
\noindent \textbf{Task:} Write a python function to check whether the given two integers have opposite sign or not.

\bigskip
\noindent \textbf{Code:}
\begin{verbatim}
def func(x,y):
    return ((x ^ y) < 0)
\end{verbatim}
\noindent \emph{Note: The function name was changed from $opposite\_Signs$ to $func$ to illustrate that the unconditional distribution lacks enough implicit information to confidently predict subsequent tokens.}

\noindent \textbf{Approximated Task:} Create a Python function that takes in two integers x and y and returns True if the signs of x and y are opposite, and False otherwise.

\begin{figure}[H]
    \centering
    \includegraphics[width=1\linewidth, trim=0.5cm 0.5cm 4.2cm 0.5cm, clip]{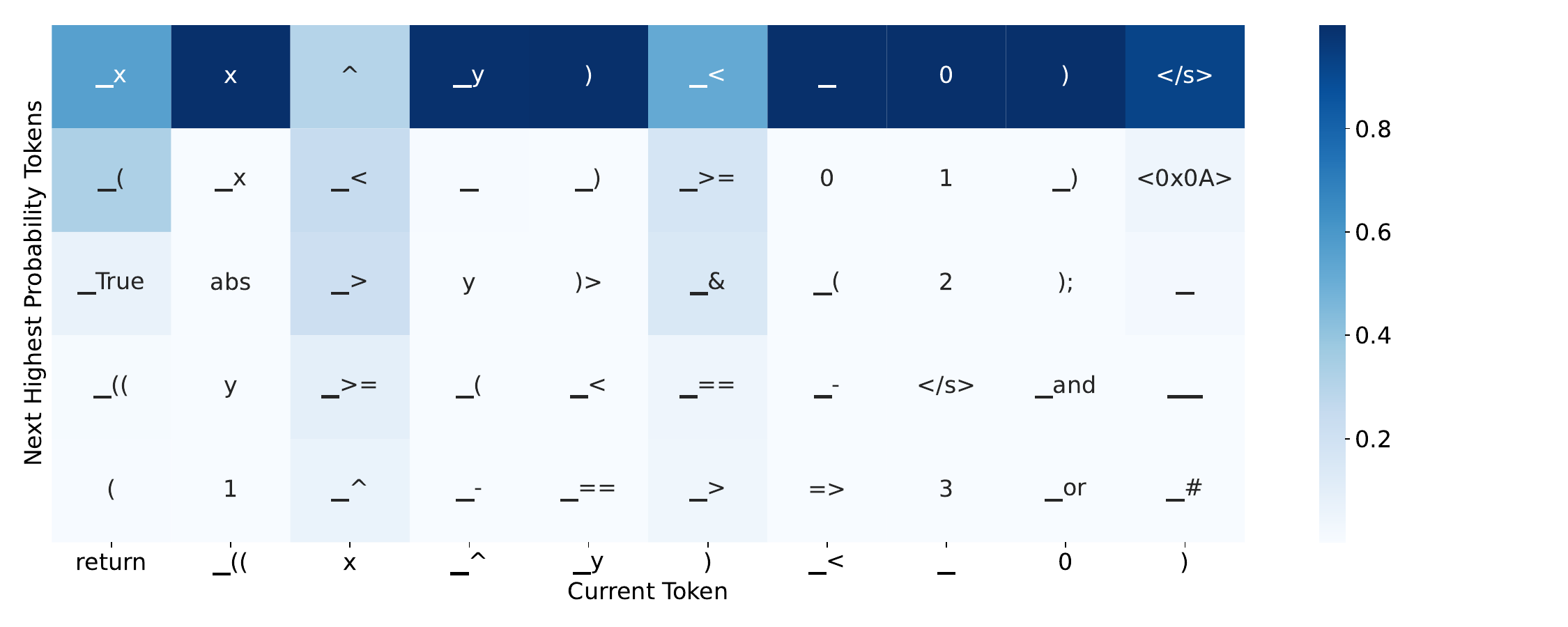}
    \caption{Heatmap of the conditional probability distribution given the \textbf{original task} for the MBPP example. Underscores and 0x0A indicate spaces and newlines, respectively.}
\end{figure}

\begin{figure}[H]
    \centering
    \includegraphics[width=1\linewidth, trim=0.5cm 0.5cm 4.2cm 0.5cm, clip]{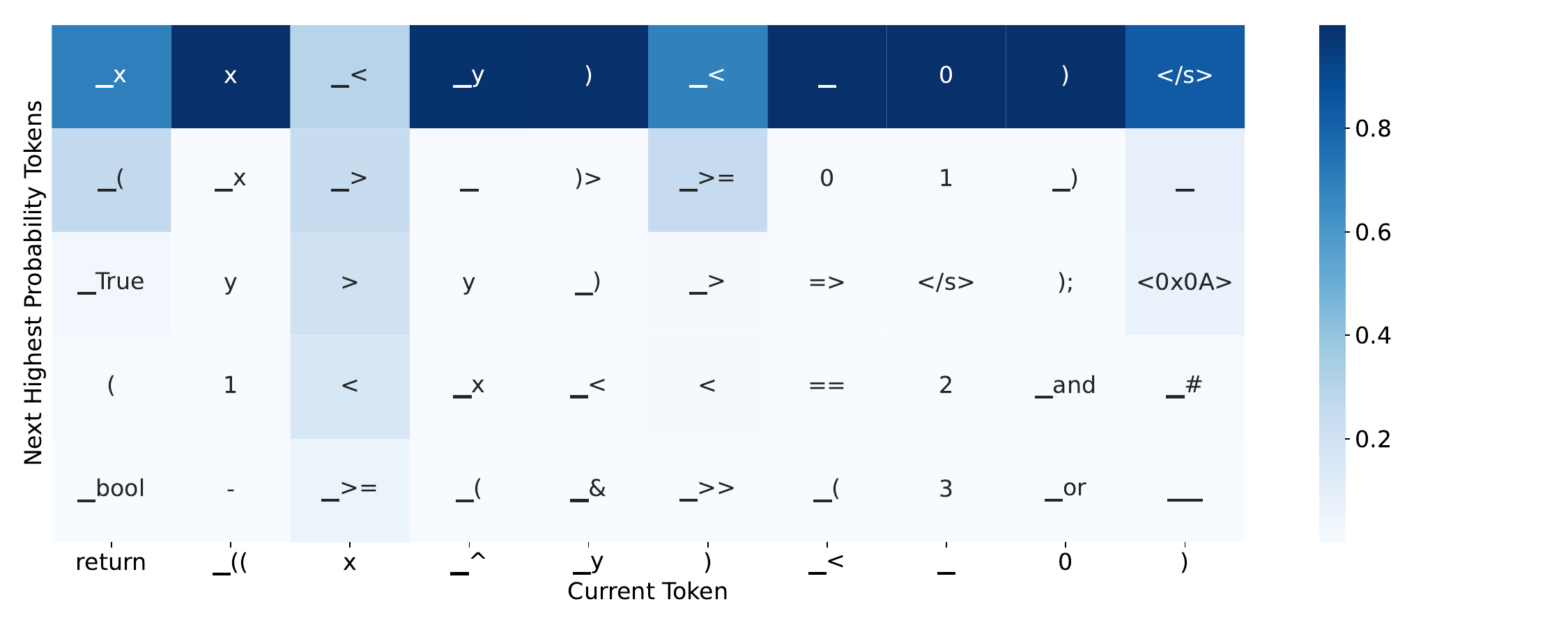}
    \caption{Heatmap of the conditional probability distribution given the \textbf{approximated task} for the MBPP example. Underscores and 0x0A indicate spaces and newlines, respectively.}
\end{figure}

\begin{figure}[H]
    \centering
    \includegraphics[width=1\linewidth, trim=0.5cm 0.5cm 4.2cm 0.5cm, clip]{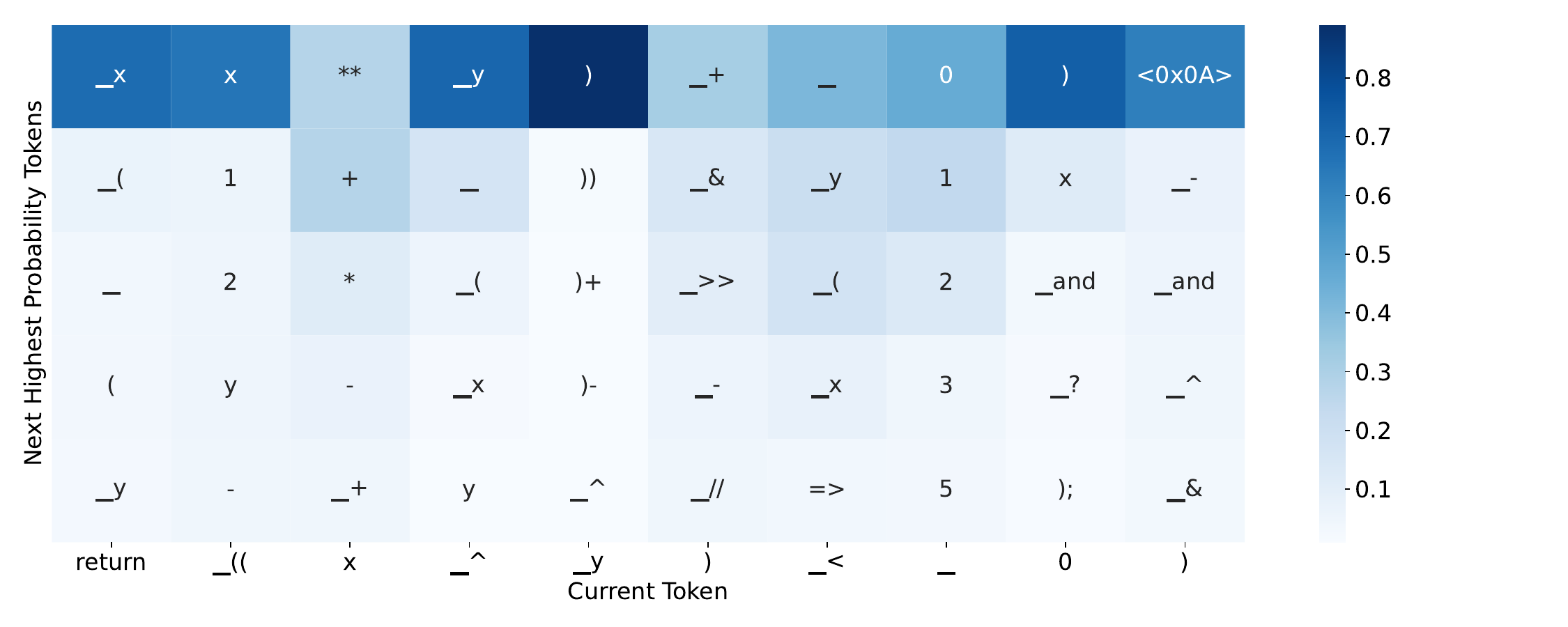}
    \caption{Heatmap of the unconditional probability distribution for the MBPP example. Underscores and 0x0A indicate spaces and newlines, respectively.}
\end{figure}

\newpage
\subsection{APPS Example}
APPS tasks are longer and more descriptive, leading to greater differences between the original and approximated versions. Despite this, the conditional probability distributions remain similar, as seen in the example below.

\bigskip
\noindent \textbf{Task:} The Rebel fleet is afraid that the Empire might want to strike back again. Princess Heidi needs to know if it is possible to assign R Rebel spaceships to guard B bases so that every base has exactly one guardian and each spaceship has exactly one assigned base (in other words, the assignment is a perfect matching). Since she knows how reckless her pilots are, she wants to be sure that any two (straight) paths – from a base to its assigned spaceship – do not intersect in the galaxy plane (that is, in 2D), and so there is no risk of collision. \emph{The task goes on to provide input and output examples, and additional notes.}

\bigskip
\noindent \textbf{Code:}
\begin{verbatim}
a, b = list(map(int, input().split()))
if a==b: print("Yes")
else: print("No")
\end{verbatim}

\noindent \textbf{Approximated Task:} Write a Python program that takes two integers as input and checks if they are equal. If they are equal, print "Yes", otherwise print "No".

\begin{figure}[H]
    \centering
    \includegraphics[width=1\linewidth, trim=0.5cm 0.5cm 4.2cm 0.5cm, clip]{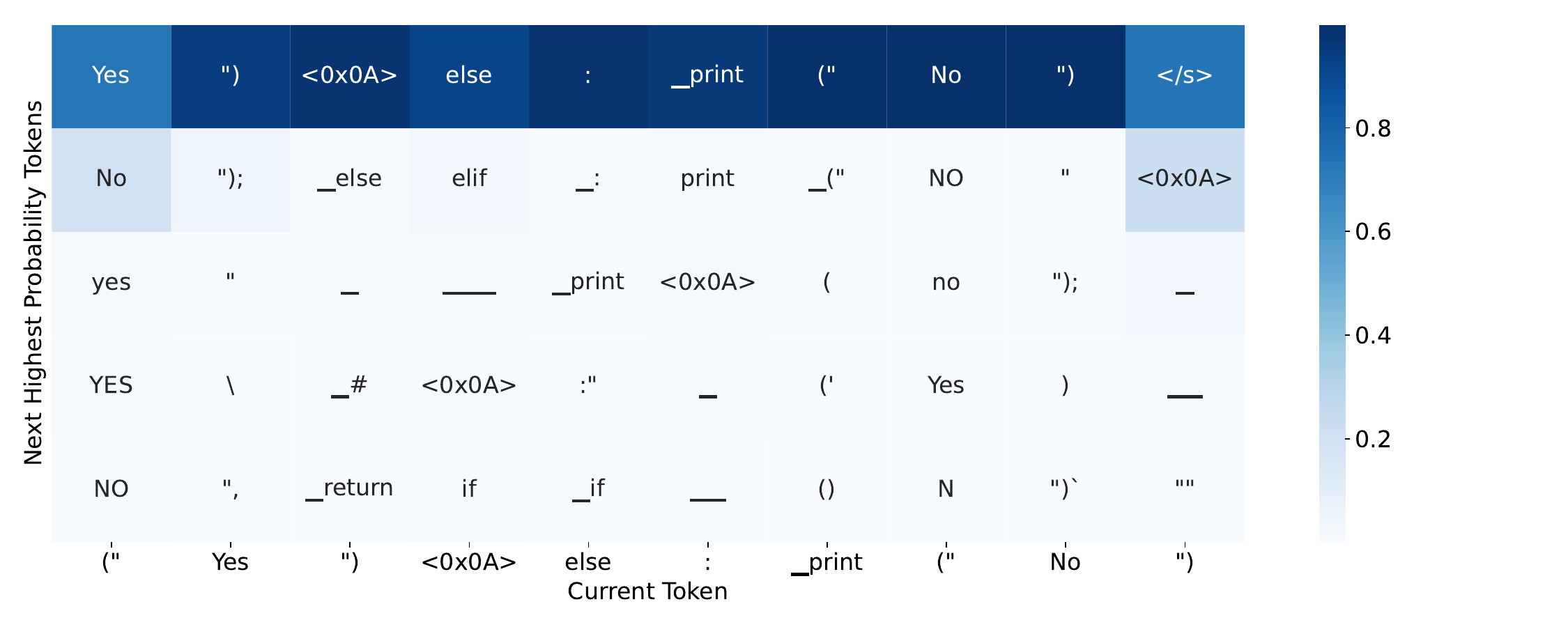}
    \caption{Heatmap of the conditional probability distribution given the \textbf{original task} for the APPS example. Underscores and 0x0A indicate spaces and newlines, respectively.}
\end{figure}

\begin{figure}[H]
    \centering
    \includegraphics[width=1\linewidth, trim=0.5cm 0.5cm 4.2cm 0.5cm, clip]{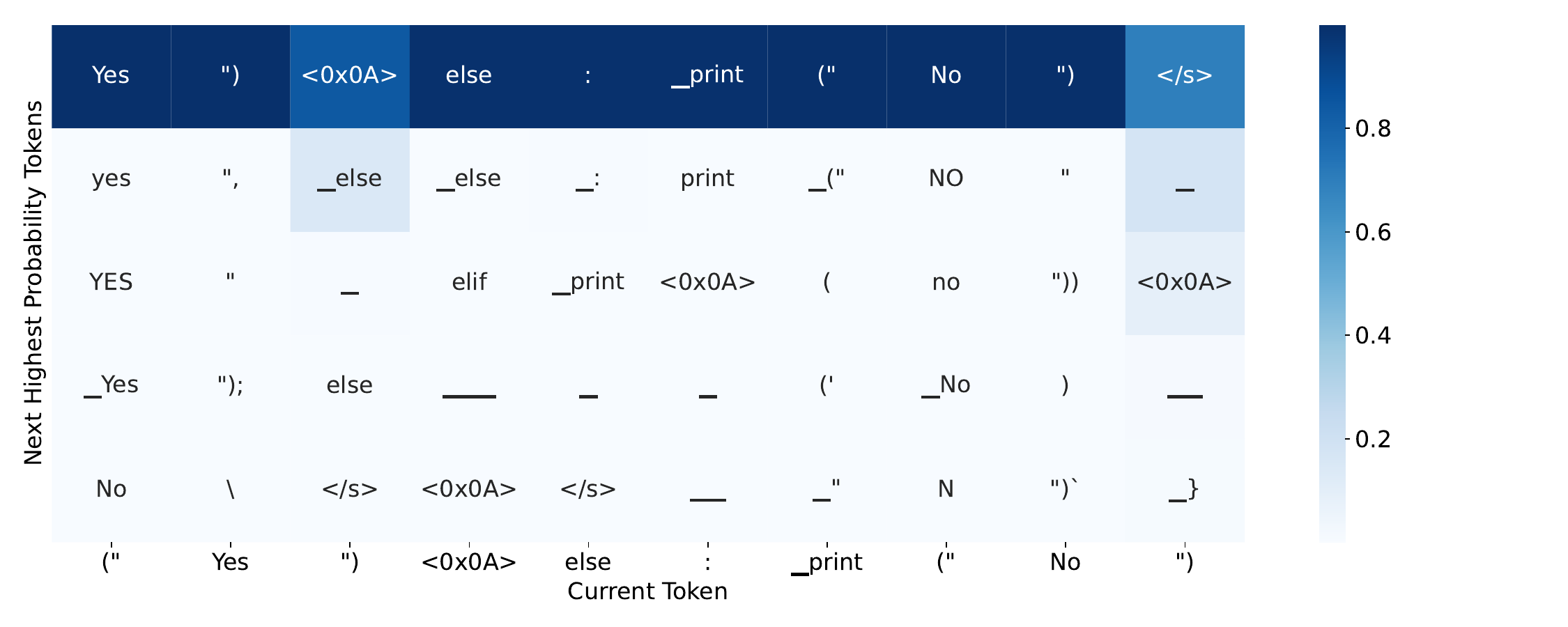}
    \caption{Heatmap of the conditional probability distribution given the \textbf{approximated task} for the APPS example. Underscores and 0x0A indicate spaces and newlines, respectively.}
\end{figure}

\begin{figure}[H]
    \centering
    \includegraphics[width=1\linewidth, trim=0.5cm 0.5cm 4.2cm 0.5cm, clip]{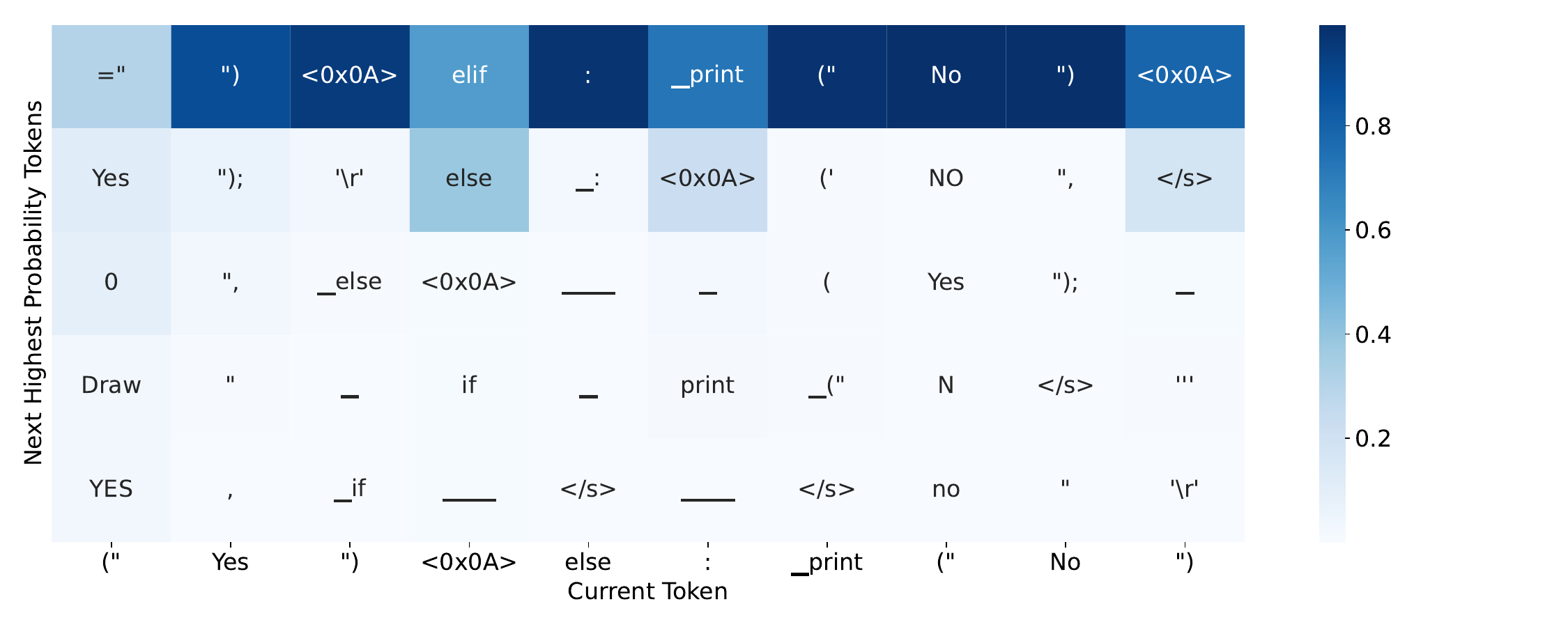}
    \caption{Heatmap of the unconditional probability distribution for the APPS example. Underscores and 0x0A indicate spaces and newlines, respectively.}
\end{figure}

\newpage
\section{Approximating Tasks with Different Seeds}
Here we provide examples of approximated tasks generated with different seeds for the same code, illustrating how using $N > 1$ averages slight variations in the conditional distribution. The examples are simple for clarity.

\subsection{MBPP Example}
\noindent\textbf{Code:}
\begin{verbatim}
def opposite_Signs(x,y): 
    return ((x ^ y) < 0);
\end{verbatim}

\noindent\textbf{Approximated Tasks:}
\begin{enumerate}
    \item Write a function that takes two arguments, \texttt{x} and \texttt{y}, and returns True if both arguments have opposite signs (positive and negative, or negative and positive), and False otherwise.
    \item Create a Python function that takes in two integers \texttt{x} and \texttt{y} and returns True if the signs of \texttt{x} and \texttt{y} are opposite, and False otherwise.
    \item Write a Python function that takes two integers \texttt{x} and \texttt{y} as input and returns a boolean value indicating whether the signs of \texttt{x} and \texttt{y} are opposite.
    \item Write a function called \texttt{opposite\_Signs} that takes two input values (\texttt{x} and \texttt{y}) and returns a boolean value indicating whether the two inputs have opposite signs (i.e., one is positive and the other is negative, or one is zero and the other is non-zero).
\end{enumerate}

\subsection{APPS Example}
\noindent\textbf{Code:}
\begin{verbatim}
a, b = list(map(int, input().split()))
if a==b: print("Yes")
else: print("No")
\end{verbatim}

\noindent\textbf{Approximated Tasks:}
\begin{enumerate}
    \item Write a Python program that takes two integers as input and checks if they are equal. If they are equal, print "Yes", otherwise print "No".
    \item Write a Python program that takes two integers as input and prints "Yes" if they are equal, and "No" otherwise.
    \item Write a Python program that reads two integers from the user, stores them in variables \texttt{a} and \texttt{b}, and then prints "Yes" if \texttt{a} is equal to \texttt{b}, and "No" otherwise.
    \item Write a Python program that takes two integers as input and checks if they are equal. If they are equal, print "Yes" otherwise print "No".
\end{enumerate}

\newpage
\section{Exploring Different Task Approximation Prompts}
\subsection{Prompts}
In this section, we examine different styles of task approximation prompts. Unless specified otherwise, all results presented in this paper use the \textit{Regular} prompt.

\subsubsection{Regular} 
\textit{System:} \textit{"You are a <LANG> developer."}  
\textit{Prompt:} \textit{"Based on the provided code snippet, create a simple one-line task description that, when given to an LLM, would likely result in the generation of a similar piece of code."}

\subsubsection{Short} 
\textit{System:} \textit{"You are a <LANG> developer."}  
\textit{Prompt:} \textit{"Based on the provided code snippet, create a very short and simple task that, when given to an LLM, would likely result in the generation of a similar piece of code."}

\subsubsection{Long} 
\textit{System:} \textit{"You are a <LANG> developer."}  
\textit{Prompt:} \textit{"Based on the provided code snippet, create a long and detailed task description that, when given to an LLM, would likely result in the generation of a similar piece of code."}

\subsubsection{Storytelling} 
\textit{System:} \textit{"You are writing a programming questions textbook. Each question is based on a short fictional story, and the reader is required to write a piece of code that solves the question in the story."}  
\textit{Prompt:} \textit{"Based on the provided code snippet, create the required story description that would likely result in the generation of a similar piece of code."}

\subsubsection{Pseudocode} 
\textit{System:} \textit{"You are a <LANG> developer experienced in writing structured pseudocode."}  
\textit{Prompt:} \textit{"Translate the given code snippet into a pseudocode-like task."}

\subsubsection{Friendly} 
\textit{System:} \textit{"You are a <LANG> developer helping a friend understand coding tasks."}  
\textit{Prompt:} \textit{"Based on the provided code snippet, create a very short and simple task that, when given to an LLM, would likely result in the generation of a similar piece of code."}

\subsubsection{Critical} 
\textit{System:} \textit{"You are a no-nonsense <LANG> developer who has no patience for inefficiency or poorly written code."}  
\textit{Prompt:} \textit{"Write a brutally honest task description that, when given to an LLM, would likely result in the generation of a similar piece of code. The tone should be direct, demanding, and critical. Do not sugarcoat anything."}

\subsection{Examples}

Here we provide examples of outputs generated using different prompt styles for task approximation, all based on the same piece of code from APPS:

\begin{verbatim}
a, b = list(map(int, input().split()))
if a == b: print("Yes")
else: print("No")
\end{verbatim}

\subsubsection{Regular}
Write a Python program that takes two integers as input and checks if they are equal. If they are equal, print "Yes", otherwise print "No".

\subsubsection{Short}
Write a Python program that prompts the user to enter two numbers, and then prints "Yes" if the numbers are equal, and "No" otherwise.

\subsubsection{Long}
Create a Python program that takes two integer inputs from the user and checks if they are equal. If the two inputs are equal, the program should print "Yes" on the console. Otherwise, it should print "No" on the console.
The program should use the \texttt{input()} function to read the input from the user, and the \texttt{split()} function to split the input into two integers. The program should then use the \texttt{map()} function to convert the input strings to integers and store them in variables \texttt{a} and \texttt{b}.
Finally, the program should use an \texttt{if} statement to check if \texttt{a} and \texttt{b} are equal. If they are equal, the program should print "Yes" on the console. Otherwise, it should print "No" on the console.
The program should be able to handle any valid input, including negative numbers and floating-point numbers. The output should be "Yes" if the two inputs are equal, and "No" otherwise.

\subsubsection{Critical}
Write a Python program that takes two integers as input and checks if they are equal. If they are equal, print "Yes" to the console. Otherwise, print "No" to the console.
\emph{Constraints:}
\begin{itemize}
    \item The input should be a space-separated list of two integers.
    \item The output should be "Yes" or "No" to the console, with no trailing or leading whitespace.
    \item The program should be efficient and use minimal resources.
    \item The code should be well-written and easy to read.
\end{itemize}

\noindent\emph{Example Input: 1 1}

\noindent\emph{Example Output: Yes}

\subsubsection{Friendly}
Given two integers, \texttt{a} and \texttt{b}, write a program to print "Yes" if they are equal, and "No" otherwise. \emph{Example:}

\begin{verbatim}
Input: 5 5
Output: Yes
Input: 5 6
Output: No
\end{verbatim}

\noindent\emph{Constraints:}
\begin{itemize}
    \item The input consists of two integers separated by a space.
    \item The input is guaranteed to be valid.
\end{itemize}

\subsubsection{Pseudocode}
Given a list of integers, check if the first and second elements are equal. If they are, print "Yes". Otherwise, print "No".

\noindent\emph{Input:} A list of integers, separated by a space. \emph{Output:} "Yes" if the first and second elements are equal, otherwise "No".

\noindent\emph{Steps:}
\begin{enumerate}
    \item Read the input list from the user.
    \item Split the input list into two elements.
    \item Convert the two elements to integers using the \texttt{int()} function.
    \item Check if the first and second elements are equal.
    \item Print "Yes" if they are equal, otherwise print "No".
\end{enumerate}

\subsubsection{Storytelling}
Alice was a developer working on a project where she needed to compare two numbers, \texttt{a} and \texttt{b}. However, she noticed that some users were inputting extra spaces or special characters at the start and end of their inputs, which were affecting the comparison. To make her code more robust, she needed a function that could strip these unnecessary characters before comparing the numbers. Help her write the relevant piece of code.

\subsection{Results on MBPP}
\begin{figure}[h]
    \centering
    \includegraphics[width=1\linewidth, trim=0.2cm 0.4cm 0.2cm 0.7cm, clip]{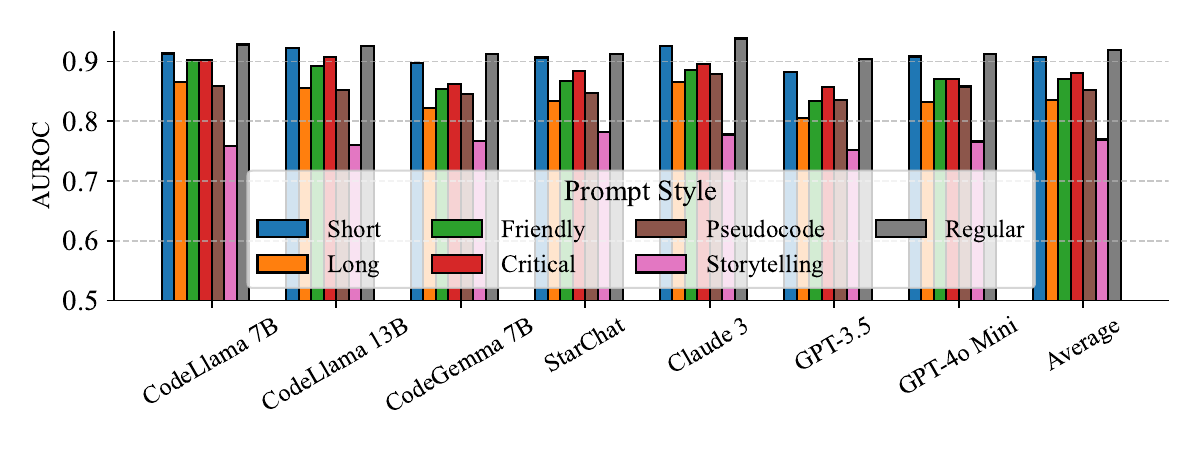}
    \caption{ATC with different prompting styles using CodeLlama7b.}
    \label{fig:prompt_styles_mbpp}
\end{figure}

\newpage
\section{Extended CPP and Java Results}
\begin{table}
\centering
\caption{Results on CodeContest with CodeLlama7b as the detector LLM.}
\label{tab:cpp_java}
\begin{tabular}{lccccccccc}
\toprule
Generator & CLlama7b & CLlama13b & Gemma & Starchat & Claude & GPT3.5 & GPT4om & Avg. \\
\midrule
\multicolumn{9}{c}{CPP} \\
\midrule
Shi \cite{shi2025detectcodegpt} & 84.05 & 81.07 & 84.05 & 73.59 & 86.03 & 81.96 & 55.31 & 78.01 \\
Ye \cite{ye2024uncovering} & - & 89.87 & - & 83.42 & - & 90.82 & - & - \\
\midrule
Entropy & 32.15 & 29.83 & 40.77 & 39.20 & 53.06 & 43.29 & 28.93 & 38.17 \\
$log P(x)$ & 71.36 & 68.80 & 74.07 & 65.23 & 78.86 & 72.11 & 48.38 & 68.40 \\
LogRank & 71.64 & 67.33 & 75.61 & 65.35 & 76.77 & 71.37 & 44.11 & 67.45 \\
LRR & 59.92 & 51.09 & 67.42 & 56.51 & 53.29 & 57.68 & 29.12 & 53.57 \\
\midrule
$ATC_{N = 1}$ & 97.03 & 97.63 & 91.18 & 90.94 & 95.63 & 92.99 & 91.06 & 93.78 \\
$ATC_{N = 2}$ & 98.00 & 98.26 & 92.87 & 91.52 & \textbf{96.71} & 93.97 & 91.51 & 94.69 \\
$ATC_{N = 4}$ & \textbf{98.28} & \textbf{98.44} & \textbf{92.99} & \textbf{91.82} & 96.62 & \textbf{94.69} & \textbf{91.62} & \textbf{94.92} \\
\midrule
\multicolumn{9}{c}{Java} \\
\midrule
Shi \cite{shi2025detectcodegpt} & 75.14 & 76.65 & 69.35 & 70.72 & 78.01 & 82.10 & 59.52 & 73.07 \\
Yang \cite{yang2023zero} & - & - & - & - & - & 64.03 & - & - \\
\midrule
Entropy & 27.28 & 32.80 & 33.87 & 33.14 & 45.33 & 41.18 & 25.43 & 34.15 \\
$log P(x)$ & 57.43 & 60.58 & 53.78 & 53.97 & 68.08 & 66.40 & 39.16 & 57.06 \\
LogRank & 49.83 & 52.50 & 45.62 & 48.98 & 59.59 & 61.61 & 31.19 & 49.90 \\
LRR & 25.64 & 21.03 & 19.04 & 28.87 & 22.94 & 34.20 & 13.74 & 23.64 \\
\midrule
$ATC_{N = 1}$ & 90.65 & 92.30 & 86.75 & 89.61 & 91.51 & 91.73 & 81.88 & 89.20 \\
$ATC_{N = 2}$ & 91.35 & 92.54 & 87.97 & 90.48 & 93.07 & \textbf{93.02} & 84.00 & 90.35 \\
$ATC_{N = 4}$ & \textbf{91.83} & \textbf{92.93} & \textbf{88.35} & \textbf{90.85} & \textbf{93.65} & 93.00 & \textbf{84.57} & \textbf{90.74} \\
\bottomrule
\end{tabular}
\end{table}

\newpage
\section{Main Results with Additional Metrics}

\begin{table}
    \centering
        \centering
        \captionof{table}{Results with Alternative Scoring Methods.}
        \label{tab:alternative_scores}
        \begin{tabular}{lcccc}
        \toprule
        Method & \multicolumn{2}{c}{Recall @ FPR 10\% Avg.} & \multicolumn{2}{c}{F1 @ Recall 90\% Avg.} \\
        \cmidrule(lr){2-3} \cmidrule(lr){4-5}
               & MBPP & APPS & MBPP & APPS \\
        \midrule
        OpenAI\textsubscript{large} & 8.15 & 17.45 & 66.63 & 66.60 \\
        \midrule
        \multicolumn{5}{c}{CodeLlama7b as Detector LLM} \\
        \midrule
        DetectGPT & 26.78 & 4.33 & 67.98 & 68.33 \\
        NPR~\cite{su2023detectllm} & 31.11 & 8.97 & 73.56 & 68.58 \\
        \midrule
        Entropy & 10.86 & 10.65 & 67.13 & 67.00 \\
        $log P(x)$ & 32.71 & 26.85 & 72.66 & 68.14  \\
        LogRank & 27.95 & 21.11 & 70.10 & 67.25  \\
        LRR~\cite{su2023detectllm} & 16.53 & 5.61 & 66.63 & 66.62 \\
        \midrule
        $ATC_{N=1}$  & 71.81 & 71.75 & 84.74 & 83.37 \\
        $ATC_{N=2}$  & 78.78 & 75.62 & 86.39 & 84.86 \\
        $ATC_{N=4}$  & \textbf{78.87} & \textbf{78.50} & \textbf{86.82} & \textbf{85.83} \\
        \midrule
        \multicolumn{5}{c}{CodeLlama13b as Detector LLM} \\
        \midrule
        Entropy & 10.09 & 9.96 & 67.11 & 66.77 \\
        $log P(x)$ & 31.23 & 24.58 & 73.28 & 67.90  \\
        LogRank & 30.16 & 21.28 & 71.12 & 67.18  \\
        LRR~\cite{su2023detectllm} & 19.18 & 7.01 & 66.67 & 66.64 \\
        \midrule
        $ATC_{N=1}$  & 76.12 & 77.72 & 86.53 & 84.16 \\
        $ATC_{N=2}$  & 80.37 & 81.58 & 88.20 & 85.86 \\
        $ATC_{N=4}$  & \textbf{83.92} & \textbf{84.08} & \textbf{89.62} & \textbf{86.96} \\
        \bottomrule
        \end{tabular}
    \hfill
\end{table}

\end{document}